\documentclass{svproc}
\usepackage[]{sidecap}
\usepackage{amsfonts}
\usepackage[dvipsnames]{xcolor}
\usepackage{amsmath}
\usepackage{dsfont}
\usepackage{url}
\usepackage{graphicx}
\usepackage{wrapfig}
\usepackage{subcaption}
\usepackage[greek,english]{babel}

\usepackage{float}
\usepackage{enumerate}
\usepackage{romannum}

\begin{document}
\mainmatter              
\title{YOLO-based Object Detection in Industry 4.0 Fischertechnik Model Environment}
\titlerunning{YOLO-based Object Detection in Fischertechnik Model Environment}  
\author{Slavomira Schneidereit\inst{1}, Ashkan Mansouri Yarahmadi\inst{2} \and
Toni Schneidereit\inst{2} \and Michael Breu{\ss}\inst{2} \and Marc Gebauer\inst{1}}
\authorrunning{Schneidereit et al.} 
\institute{
BTU Cottbus-Senftenberg, Chair of Automation technology, Platz der Deutschen Einheit 1, 03046 Cottbus, Germany\\
\email{Slavomira.Galovicova,gebaumar@b-tu.de},
\and 
BTU Cottbus-Senftenberg, Institute for Mathematics, Platz der Deutschen Einheit 1, 03046 Cottbus, Germany\\
\email{yarahmadi,breuss,schneton@b-tu.de},
}

\maketitle              

\begin{abstract}
In this paper we extensively explore the suitability of YOLO architectures to monitor the process flow across a Fischertechnik industry 4.0 application. Specifically, different YOLO architectures in terms of size and complexity design along with different prior-shapes assignment strategies are adopted. To simulate the real world factory environment, we prepared a rich dataset augmented with different distortions that highly enhance and in some cases degrade our image qualities. The degradation is performed to account for environmental variations and enhancements opt to compensate the color correlations that we face while preparing our dataset. The analysis of our conducted experiments shows the effectiveness of the presented approach evaluated using different measures along with the training and validation strategies that we tailored to tackle the unavoidable color correlations that the problem at hand inherits by nature.
\end{abstract}
\section{Introduction}
\label{sec:Introduction}
%
Within the context of computer vision, the task of simultaneously performing object localisation and recognition in an given image is perhaps one of the most interesting and important tasks with a broad range of applications on different environments, namely underwater, on the ground and in the sky~\cite{BTAAK2018,WY2021,ZR2019}. In general, there are mainly two types of state-of-the-art object detectors, namely those based on only a single or two-stage detection phases. One can refer to methods proposed in~\cite{GDDM2013} and~\cite{LAES2016,YOLOv1} as the most well known representatives for each type and also to~\cite{SSAMSetAl2021} for a concrete survey on the state-of-the-art deep learning based object detection approaches.

Let us refer to the line of R-CNN work devised in~\cite{GDDM2013}, that enriches a set of extracted overlapping region proposals from the input image using CNN and performs the classification task on them by adopting a linear SVM~\cite{CV1995}. R-CNN works according to the motto~\emph{recognition using regions} introduced in~\cite{CLAM2009}. The Faster R-CNN~\cite{RHGS2015} reported the R-CNN region proposal computation to be the running time bottle-neck and substituted it by a Region Proposal Network (RPN) as a fully convolutional network that simultaneously predicts object bounds and their scores at each position and fed them to the detection network. The so called Mask R-CNN~\cite{HGDG2017} was an improvement over Faster R-CNN, by adding a branch for predicting segmentation masks on each region proposal and in parallel with the existing branch for classification and region proposal regression. The mask branch is a small fully connected layer (FCN) applied to each region proposal, predicting a segmentation mask in a pixel-to-pixel manner.

In contrast to the R-CNNs~\cite{GDDM2013,HGDG2017,RHGS2015}, YOLOv1~\cite{YOLOv1} is extremely faster since the detection task is framed in general as a regression problem, in a sense that a single convolutional network simultaneously predicts multiple bounding boxes and class probabilities for those boxes. Here, the regional proposal computation, as a distinct property of R-CNN based approaches, is avoided to achieve the state-of-the-art run time efficiency. In addition, YOLOv1 reasons globally about the entire image and all objects appeared in it, that means it simultaneously predicts each bounding box and assigns its corresponding label across all classes. This property has root in model architecture of the YOLOv1 that will be explained later. However, YOLOv1 has a relatively lower recall rate and localisation accuracy compared to proposal-based methods, that was compensated in YOLOv2~\cite{YOLOv2} and its successors YOLOv3~\cite{YOLOv3}, YOLOv4~\cite{YOLOv4} and YOLOv5~\cite{YOLOv5}. One can refer to~\cite{BTAAK2018,MZLXX2020,ZR2019} and~\cite{JJS2020} that report YOLOv3 and YOLOv4, respectively, to have better mAP measures~\cite{EEGWWZ2015} compared to Faster R-CNN algorithms.

Nevertheless, the recent major advances in object detection are still expected to be integrated within the emerging family of YOLO. Among them, one can refer to anchor-free detection mechanism~\cite{HJ2018} as all of the YOLO versions addressed in current study rely on anchors as prior-bounding boxes with predefined ratios between their heights and widths. Adoption of an anchor-free mechanism significantly reduces the number of model parameters and avoids the task of heuristic tuning to obtain initial prior-boxes~\cite{YOLOv2}. As an anchor-free approach authors of~\cite{HJ2018} introduced the CornerNet as a single-stage object detection approach to eliminate the critical role of anchor boxes within the object detection context. The so called corner pooling layer as a new type of pooling layer was introduced as part of the CornerNet to localize corners of bounding boxes. In this way, the pre-processing task of prior-box creation can be replaced by the convolutional neural network CornerNet introduced in~\cite{HJ2018}.

In current study, we will closely discuss different aspects of YOLOv1~\cite{YOLOv1}, concerning model architecture, the cost function and the way in which YOLOv1 realises the single-stage~\cite{WDDCSCA2015} object detection principle. Later, we feature important improvements appeared in more advanced versions, namely YOLOv3~\cite{YOLOv3} and YOLOv5~\cite{YOLOv5}. In addition, YOLOv5 consists of four versions on its own, which are YOLOv5s, YOLOv5m, YOLOv5l, and YOLOv5x. This is classified according to the memory storage size, but the principle is the
the same~\cite{HG2022}. In a similar ways, YOLOv3 is also developed in two small and large scales, YOLOv3s and YOLOv3l respectively. In current study, we report our results based on YOLOv3s, YOLOv3l, YOLOv5s and YOLOv5l.

In specific, our major contributions consist of investigate the applicability of the latest development technology in the field of YOLO object detectors to localise and classify our objects of interest transported across an industry $4.0$ manufacturing simulator shown as Fig.~\ref{fig:FTF}. Our simulator is further augmented with a wide range of spatial image processing task~\cite{AetAl2020}, after being recorded by a Logi HD $1080$p webcam. In Sub-section~\ref{sec:Dataset}, we clearly explain the creation of our augmented dataset to account for any environmental variation that may occur in real world scenarios. 
\section{The Fischertechnik Learning Factory}
\label{sec:FTLF}
\begin{figure}
  \begin{minipage}[c]{0.5\textwidth}
    \includegraphics[width=.85\textwidth]{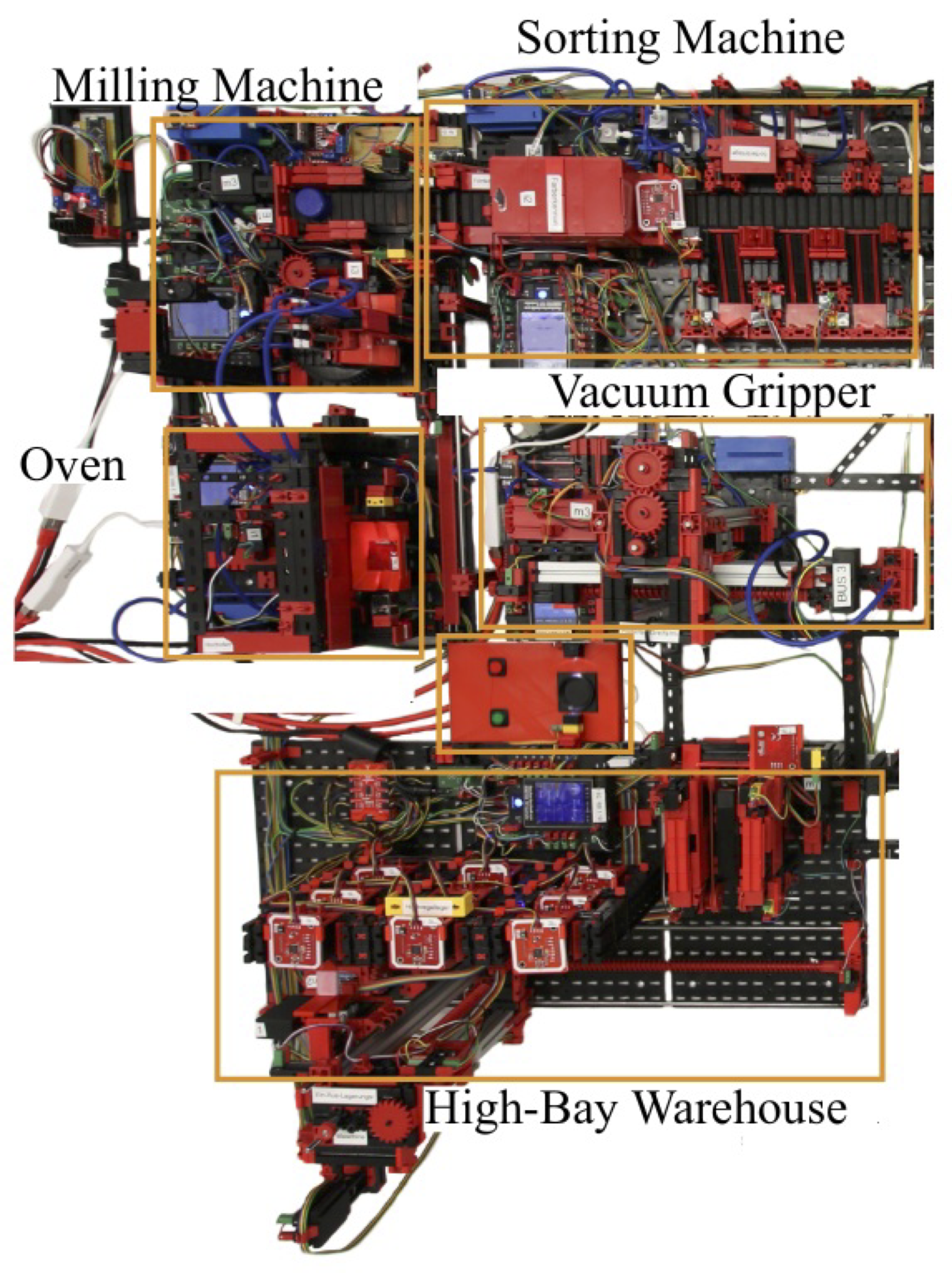}
  \end{minipage}\hfill
  \begin{minipage}[c]{0.5\textwidth}
   \caption{The used factory model consists of five individual modules, namely the sorting machine, the oven, the milling machine, the high-bay warehouse and a vacuum gripper~\cite{JLetAl2022}.}
   \label{fig:FTF}
  \end{minipage}
\end{figure}

In current study we used a \textit{Fischertechnik} factory model (See Fig.~\ref{fig:FTF}) to simulate an Industry $4.0$ manufacturing environment. Such models are also known as Learning Factories~\cite{ACJetAl2017} and adopted within the context of industry 4.0 research and training purposes~\cite{KB2019}. 

\begin{figure}
  \begin{minipage}[c]{0.5\textwidth}
    \includegraphics[width=.85\textwidth]{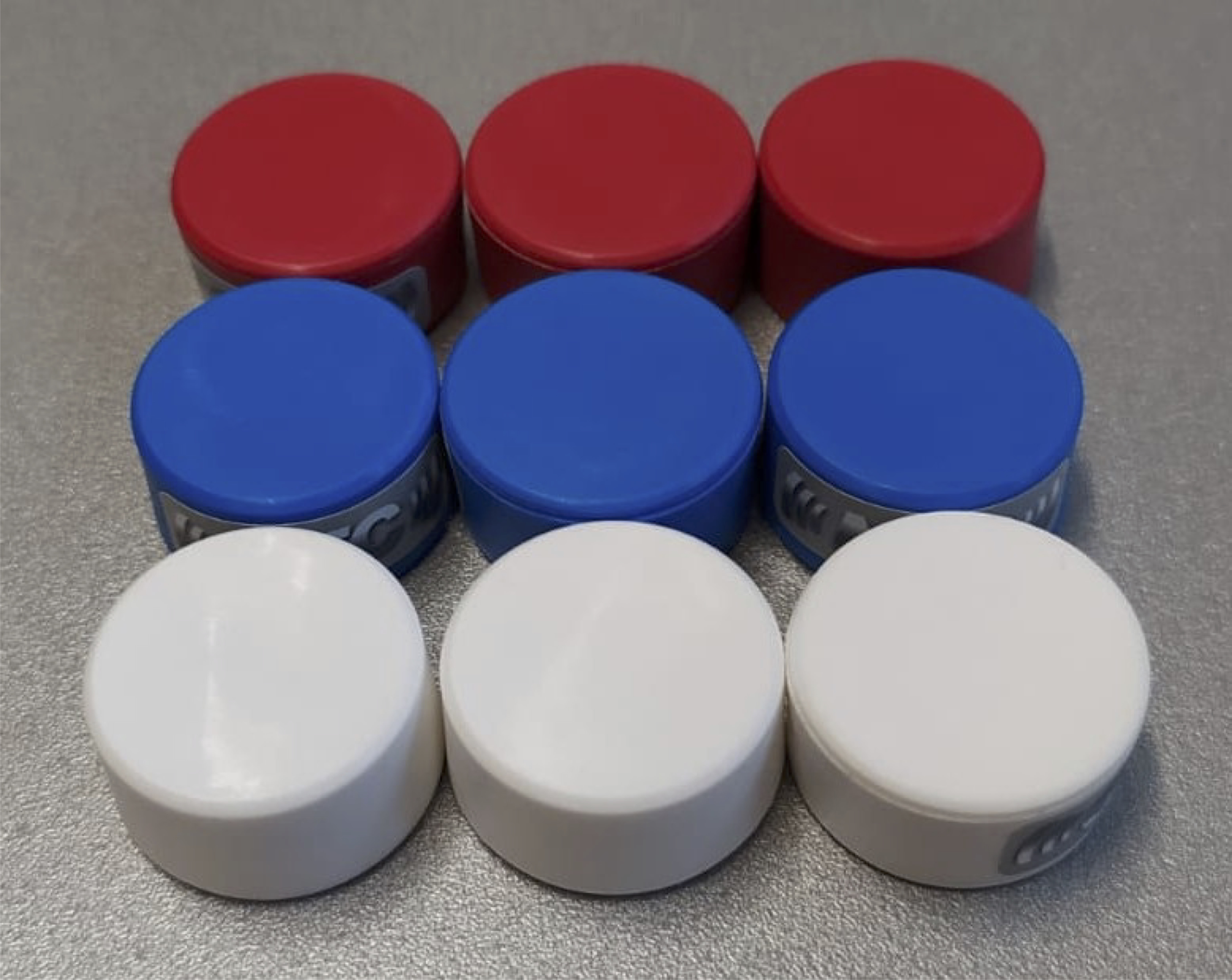}
  \end{minipage}\hfill
  \begin{minipage}[c]{0.5\textwidth}
    \caption{The work pieces corresponding to the factory that are transported across the modules and recorded by a Logi HD $1080$p webcam used to establish our data set. As one observed there is a very high correlation among the color of the red pieces and the factory itself shown as Fig.~\ref{fig:FTF}. Such a correlation is also observed partially concerning the blue color.}
  \label{fig:Pieces}
  \end{minipage}
\end{figure}

The adopted model by us consists of five individual modules, namely the sorting machine, the oven, the milling machine, the high-bay warehouse and a vacuum gripper. Each module of the factory works independently and is steered by its own controller that communicates and synchronises itself with other controllers via an Ethernet network. The working pipeline of the factory starts by taking the accumulated work pieces (See Fig.~\ref{fig:Pieces}) from the high-bay warehouse and move them to the oven and the milling machine later sorted and finally transported by the vacuum gripper and stored in the starting high-bay warehouse~\cite{JLetAl2022}. 
%
\section{Dataset Preparation} 
\label{sec:Dataset}
%
Prior to the training of the YOLO models, a data set consisting of $700$ images of the factory is captured showing the work pieces with different colors being transported across it. The images are divided into seven batches each containing $100$ number of images. Each work piece with a particular color appears exactly $400$ times and distributed across four different batches such that each batch contains at least one and at most three different color work pieces. In this way a fair distribution of colors across batches is guaranteed aiming to avoid any probable color bias effect during the training process~\cite{NFNKA2019}. A tabular representation of our dataset creation strategy is shown as Table~\ref{data_batches} with $\times$ and $-$ symbols used to reveal the presence or absence of a particular color within each of the seven batches $b_i$ with $i\in\left\{1,2,\ldots,7\right\}$. More specific, the $\times$ symbol indicates that one work piece of the corresponding color is visible in each image of the batch. For example, each image in batch $b_1$ contains one white work piece, while each image in batch $b_7$ has one white, one blue and one red work piece visible. Therefore, number of labeled work pieces that is $1200$ is more than the total number of images, namely $700$.   
\begin{table}[!h]
\centering
\caption{\label{data_batches} A tabular representation of our dataset creation strategy. Each color contributes to the creation of exactly four number of batches $b_i$ and each batch consists of one, two or in one case three different colors, namely $b_7$. Here $\times$ and $-$ symbols are used to reveal the presence or absence of a particular color within each of the batches. Note that, on each row $400$ work pieces with a particular color are distributed across their corresponding batches with each batch to contain $100$ number of images. In total, we have $700$ number of images containing $1200$ number of labeled color pieces.}
\begin{tabular}{c c c c c c c c c}
 & $b_1$ & $b_2$ & $b_3$ & $b_4$ & $b_5$ & $b_6$ & $b_7$  \\
\hline
White & $\times$ & $-$ & $-$ & $-$ & $\times$ & $\times$ & $\times$  \\
\hline
Blue  & $-$ & $\times$ & $-$ & $\times$ & $-$ & $\times$ & $\times$  \\
\hline
Red  & $-$ & $-$ & $\times$ & $\times$ & $\times$ & $-$ & $\times$  \\
\hline
\end{tabular}
\label{tab:TableImage}
\end{table}

However, the captured images do not necessarily represent the characteristics of real environmental variations. Hence, we opt to perform a data augmentation phase in direction of increasing the variability of our images aiming to have better work piece detection results. In particular we took six different pixel-wise operations based on the toolbox~\cite{AetAl2020}, namely Gaussian blur, brightness, linear contrast, motion blue, noise and sharpness adjustments to augment our initial set of created images. This consequently means the original set of $700$ number of images gathered based on the strategy explained in Table~\ref{tab:TableImage} is put to gather along with its six augmented versions created based on the six pixel-wise adjustments resulting in total $4900$ number of augmented images comprising our desired dataset that we use for the training purpose. The augmentation detailed parameters used are $\sigma_1\in\{1,2,3,4\}$ as the standard deviation of the Gaussian kernel to blur the image, $k_1\in\{0.5,0.7,1.3,1.5\}$ as the coefficients multiplied to pixel values to increase their brightness, $k_2\in\{0.5,0.7,1.3,1.5\}$ to linearly improve the contrast of the pixels, with a bigger $k_2$ resulting to more contrast enhancement. To simulate the blur because of the camera movement, namely motion blue, kernels with sizes $5\times 5$, $10\times 10$, $15\times 15$ and $20\times 20$ are used while additional Gaussian noise is added with standard deviations of $\sigma_2\in\{0.05,0.1,0.2,0.25,0.3\}$ to impact the severity of the added noise. In general, the bigger are the kernel size, $\sigma_1$ and $\sigma_2$, the more sever is their corresponding augmentations in terms of changing the image quality. Finally, the sharpness of our images are adjusted by adopting $k_3\in\{0.0,0.25,0.5,1.5,2.0\}$ with values less than $1.0$ to decrease and bigger than $1.0$ to increase the sharpness.
%
\section{YOLO Models}
\label{sec:YOLOModels}
In the following, we elaborate the model architectures of YOLOv1~\cite{YOLOv1} along with YOLOv3~\cite{YOLOv3} and YOLOv5~\cite{YOLOv5}.
%
\subsection{YOLOv1}
\label{subsec:YOLO-1}
The YOLOv1 design enables end-to-end training by dividing the input image into a set of $S \times S$ grid cells that are convoluted to obtain spatially extracted features passed to fully connected layers to predict the output confidence scores and coordinates of the enclosed bounding boxes to the objects. YOLOv1 computes two confidence scores, namely the object and the class confidence score that we explain as follow. 

Within each grid cell, $B=2$ number of bounding boxes are found and their confidence scores are formally defined as $P\left(o\right)\cdot IOU$, with $P\left(\cdot\right)$ to be the objectness probability. More precisely, the object confidence score represents the amount of the objectness, namely how likely a found box contains an object $o$, and how accurate is the position of the boundary box. In practice, the box confidence score is calculated~\cite{YOLOv1} based on IOU, to be the intersection over union metric~\cite{RTGRS2019}, concerning the predicted and ground truth boxes of the object $o$.

In addition, a set of $s\in\mathbb{N}$ number of conditional class probabilities $P\left(\kappa_s \mid o\right)$ concerning each grid cell is computed. Such probabilities matter when the centre of the object $o$ locates in a grid cell and reveal the likeliness of the found object in a cell to fall under each of the $\kappa_s$ classes. This consequently means, if a grid cell has boxes with low object confidence scores $P\left(o\right)\cdot IOU$, the lower will be the probability of that cell being responsible for any class. The class confidence score reads
\begin{equation}
  \label{eq:CCS}
  P\left(\kappa_s\right)
  = P\left(\kappa_s \mid o\right)\cdot \left(P\left(o\right) \cdot IOU\right)
\end{equation}
Here, a small confidence score $P\left(o\right)\cdot IOU$ results in a small class confidence score~\eqref{eq:CCS}. This holds also true in case IOU is small and near to zero. 

\begin{figure}
\centering
\includegraphics[width=\textwidth]{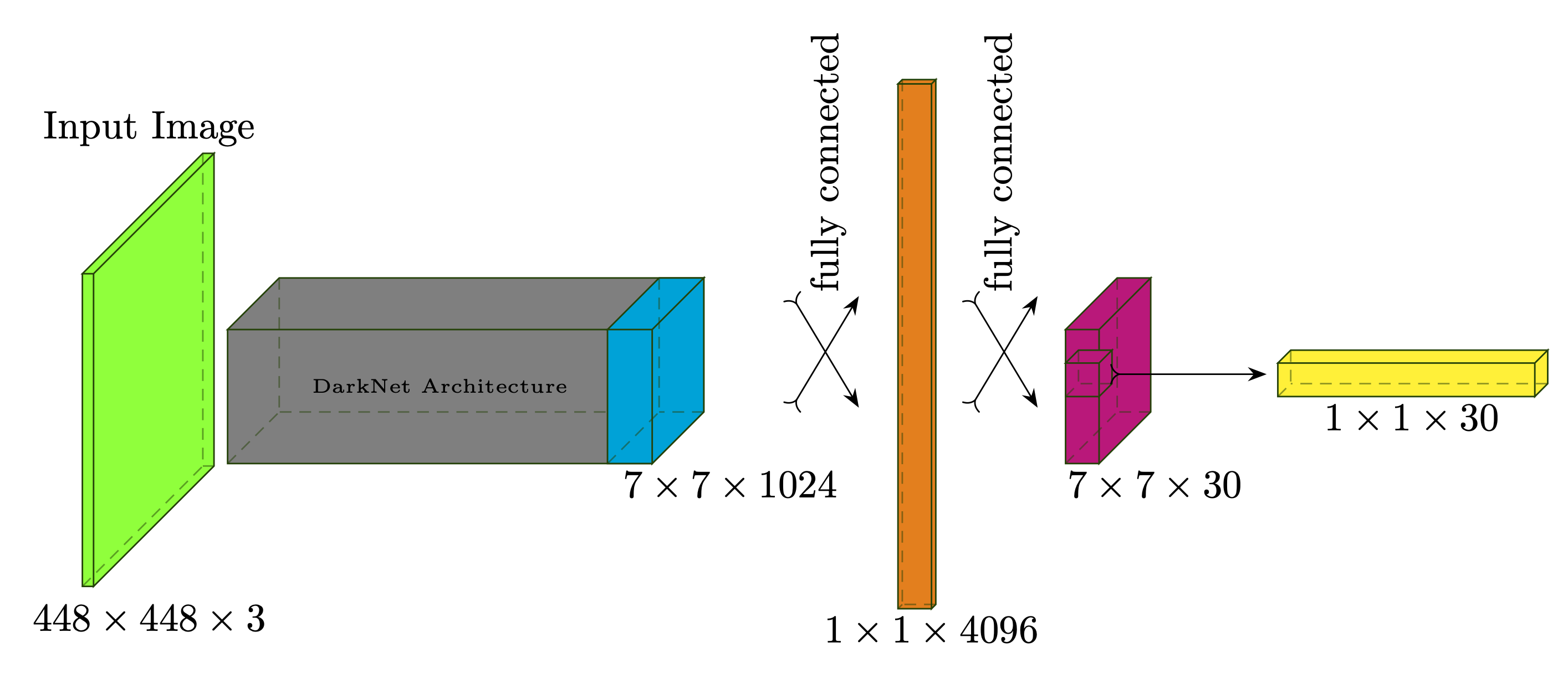}
\caption{A schematic view of the YOLOv1 developed using DarkNet framework~\cite{darknet}. The slice of size $1\times 1\times 30$ as a subset of the last YOLOv1 layer corresponds to a cell from input image. It comprised of $20$ number of conditional probabilities along with two predictor subsets of sizes $1\times 1\times 5$ containing the $x$ and $y$, as the centroid, $h$ and $w$ as the height and width and IOU score corresponding to two found boxes within each grid cell of the input image.}
\label{fig:YOLOv1-arch}
\end{figure}

Let us start investigating the YOLOv1 model architecture (See Fig.~\ref{fig:YOLOv1-arch}) from the last layer, comprised of $7\times 7\times 30$ number of tensor predictors. As YOLOv1 network architecture is inspired by the GoogLeNet model~\cite{SLJSRAEVR2014}, with $24$ convolutional layers, that down samples the input training images of size $448\times 448\times 3$ into $7\times 7\times 30$. A closer look to a slice of size $1\times 1\times 30$ as a subset of the last YOLOv1 layer, that corresponds to a cell from input image, reveals the value of $30$ to be comprised of $20$ number of conditional probabilities $P\left(\kappa_s \mid o\right)$, as YOLOv1 uses the PASCAL Visual Object Classes Challenge (VOC)~\cite{EEGWWZ2015} with $20$ labelled classes, along with two predictor subsets of sizes $1\times 1\times 5$ containing the $x$ and $y$, as the centroid, $h$ and $w$ as the height and width and IOU score~\cite{RTGRS2019} corresponding to two found boxes within each grid cell. The width and height of bounding boxes are normalized by their corresponding image width and height~\cite{YOLOv1}, so that they fall between $0$ and $1$. In addition as reported by~\cite{YOLOv1}, the bounding box coordinate centres $x$ and $y$ are also parameterized to be offsets of their corresponding grid cell locations so they are also bounded between $0$ and $1$.

YOLOv1 unifies~\cite{YOLOv1} the separate components of object detection along with classification task based on a composed loss function from three parts, namely the confidence loss, the localisation loss and the classification loss.

The localisation loss measures the errors in the predicted boundary box locations and sizes. Within each cell, only the box responsible for detecting the object is considered while the localisation loss is computed. The ``responsible box'' among all $B$ number of predicted boxes is the one with the highest current IOU with the ground truth during the training stage. In addition, the localisation loss concerns about small deviations depending if they occur in small or large boxes. This is to stress that while computing the IOU, small deviation in large boxes matter less than the same deviation in small boxes and is partially addresses by embedding the square roots of the bounding box width and height instead of the width and height as part of the localisation loss. With this, the localisation loss reads
\begin{equation} \label{eq:ll}
    \begin{split}    
    \mathcal{L}_l\left(x,\hat{x},y,\hat{y},\lambda_{coord}\right) 
    & = \lambda_{coord} \sum_{i=0}^{S^2} \sum_{j=0}^{B} \mathds{1}^{obj}_{ij} \left[\left(x_i-\hat{x}_i\right)^2+\left(y_i-\hat{y}_i\right)^2\right] \\
    & + \lambda_{coord} \sum_{i=0}^{S^2} \sum_{j=0}^{B} \mathds{1}^{obj}_{ij} \left[\left(\sqrt{w_i}-\sqrt{\hat{w}_i}\right)^2+\left(\sqrt{h_i}-\sqrt{\hat{h}_i}\right)^2\right] 
    \end{split}
\end{equation}
with $\mathds{1}^{obj}_{ij}$ to be $1$ if the $j$ bounding box in $i$ cell is responsible for detecting the object, otherwise $0$. 

The coefficient $\lambda_{coord}$ used within the localisation loss~\eqref{eq:ll} can be well explained in contrast to the $\lambda_{noobj}$ used in confidence loss~\eqref{eq:cl} as their usage is to avoid model instability during the training stage. In general, many grid cells within each training image represent the background and, as the training proceeds, their ``object confidence'' score predictors~\cite{RTGRS2019} are pushed towards zero though they were falsely considered as box candidates. This alternatively leads to an overpower gradient emerging from the cells containing objects with their ``object confidence'' score predictors encouraged towards $1$. This unbiased gradient results to an early instability during the training phase. Authors of~\cite{YOLOv1} resolved this issue by reducing the loss corresponding to those falsely accepted background bounding boxes by letting $\lambda_{noobj}$ in~\eqref{eq:cl} to have a relatively smaller number $0.5$ compared to a larger value $5$ assigned in~\eqref{eq:ll} to $\lambda_{coord}$ that still encourages YOLO to find bounding boxes truly representing the objects. The confidence loss reads
\begin{equation} \label{eq:cl}
    \begin{split}    
    \mathcal{L}_c\left(C_i,\hat{C}_i,\lambda_{noobj}\right)     
    & = \sum_{i=0}^{S^2} \sum_{j=0}^{B} \mathds{1}^{obj}_{ij} (C_i-\hat{C}_i)^2
     + \lambda_{noobj} \sum_{i=0}^{S^2} \sum_{j=0}^{B} \mathds{1}^{noobj}_{ij} (C_i-\hat{C}_i)^2 \\
    \end{split}
\end{equation}
with $\mathds{1}^{noobj}_{ij}$ to be $1$ if the $j$ bounding box in $i$ cell contains no detected object, otherwise $0$. Here, $C$ and $\hat{C}$ both in $\left[0,1\right]$ are the ground truth and objectness scores. In case an object appears in a ground truth box then $\hat{C}=1$ and otherwise equal to $0$. The IOU computed between the predicted box and ground truth box is considered as $\hat{C}$.

Next YOLOv1 opts to estimate a classification loss
\begin{equation} \label{eq:al}
    \begin{split} 
    \mathcal{L}_a\left(P_i\left(\kappa_s\right),\hat{P}_i\left(\kappa_s\right)\right) 
    & = \sum_{i=0}^{S^2} \mathds{1}^{obj}_{i} \sum_{\kappa \in \text{classes}} \left(P_i\left(\kappa_s\right)-\hat{P}_i\left(\kappa_s\right)\right)^2
    \end{split}
\end{equation}
with $P_i\left(\kappa_s\right)$ obtained from~\eqref{eq:CCS} to represent the conditional probabilities of a detected object $o$ in a cell to fall under each of the classes $\kappa_s$. To this end, the YOLOv1 total loss is deduced in~\cite{YOLOv1} as
\begin{equation} \label{eq:tl}
\mathcal{L} = \mathcal{L}_a+\mathcal{L}_c+\mathcal{L}_l.
\end{equation}
%
\subsection{YOLOv3}
\label{subsec:YOLO-3}
A distinct advantage of YOLOv3 compared to YOLOv1 is its multi scale prediction capability. This is achieved by adopting Darknet-$53$~\cite{darknet}, which originally has $53$ layer network trained on ImageNet~\cite{ImageNet}. For detection task, $53$ more layers are stacked on top of it, giving us a $106$ layer fully convolutional underlying architecture for YOLOv3. Across the stacked-Darknet with $106$ layers, the up-sampling and concatenation methods are used three times where feature maps of sizes $13\times 13 \times 255$, $26\times 26 \times 255$ and $52\times 52 \times 255$ are produced. As reported in~\cite{YOLOv3} and to produce these feature maps, their corresponding feature maps from two previous layers are up-sampled by scale of $4$ and then concatenated with their corresponding earlier feature maps from the network. As authors of~\cite{YOLOv3} claim, adoption of this technique, schematically shown as Fig.~\ref{fig:FPN}, called Feature Pyramid Network (FPN)~\cite{LDGHHB2016} is to obtain meaningful semantic information from the up-sampled features and finer-grained information from the earlier feature maps. It is noteworthy that, in case of dealing with scaled $13\times 13$, $26\times 26$ or $52\times 52$ feature maps to detect on different scales, the located grid cells on input image also need to have the same dimensionality so that a one to one correspondence among the cell numbers within the input image and the scaled feature maps hold true. In this case, each grid cell within the input image still stays responsible to account for an object centroid that it contains, and at the last layer its corresponding $1\times 1\times 255$ predictor tensor will be interpreted to find the suitable bounding box and class label. 

In YOLOv3, each output tensor of size $1\times 1\times 255$ comprises $B=3$ number of bounding boxes each represented with six attributes, namely the centroid coordinates, the dimensions, the objectness score and the last one to be a set of $80$~\cite{LMBBGHPRDZ2014} number of conditional class confidences. As YOLOv3 predicts on three different scales, in total nine number of ``derived'' bounding boxes are predicted. The derivation is in fact performed from a set of the so called ``anchor boxes'' that already provided to YOLOv3 by the ``dimension clusters'' pre-processing stage~\cite{YOLOv2}. 

The main motivation behind the anchor boxes is to have a limited set of prior-shapes deduced from the dataset and the ground truth boxes at hand, so that during the training phase the ground truth boxes are compared against them and a transformation between the prior and ground truth boxes are learnt. Here, the anchor box that has the highest IOU with the ground truth box will be chosen to train the model. A K-means clustering~\cite{L1982} approach selects in total nine anchor boxes on different scales as the mean anchor box of each of the nine established clusters on different scales and with respect to the COCO dataset~\cite{LMBBGHPRDZ2014}. The main advantage of prior boxes is to make YOLOv3 capable of predicting multiple objects with their shapes to have different height and width aspect ratios on varying scales. The desired learnt transformation to the anchor boxes is applied as
\begin{equation} \label{eq:tr}
\begin{split}
b_x&=\sigma\left(t_x\right)+c_x\\
b_y&=\sigma\left(t_y\right)+c_y\\
b_w&=p_w\cdot e^{t_w}\\
b_h&=p_h\cdot e^{t_h}    
\end{split}
\end{equation}
with $b_x$, $b_y$, $b_w$, $b_h$ are the coordinate centre, width and height of the final predicted box after the transformation along with $t_x$, $t_y$, $t_w$, $t_h$ to be the the predicted values of the box by model. The $c_x$ and $c_y$ are the top-left coordinates of the corresponding grid cell and $p_w$ and $p_h$ are the anchor box dimensions. Note that, in some cases the model produces centroid coordinates $c_x$ and $c_y$ to be located in neighbor cells, that are brought back to the current cell by applying a sigmoid function $\sigma$. Also the exponential function is used to correct the obtained dimensions $t_h$ and $t_w$ of the box if they are produced as negative values by the model, so that they can be transformed to the suitable anchor box during the training phase.
\subsection{YOLOv5}
\label{subsec:YOLO-5}
\begin{figure}
\centering
\begin{subfigure}[b]{0.44\textwidth}
\centering
\includegraphics[width=\textwidth]{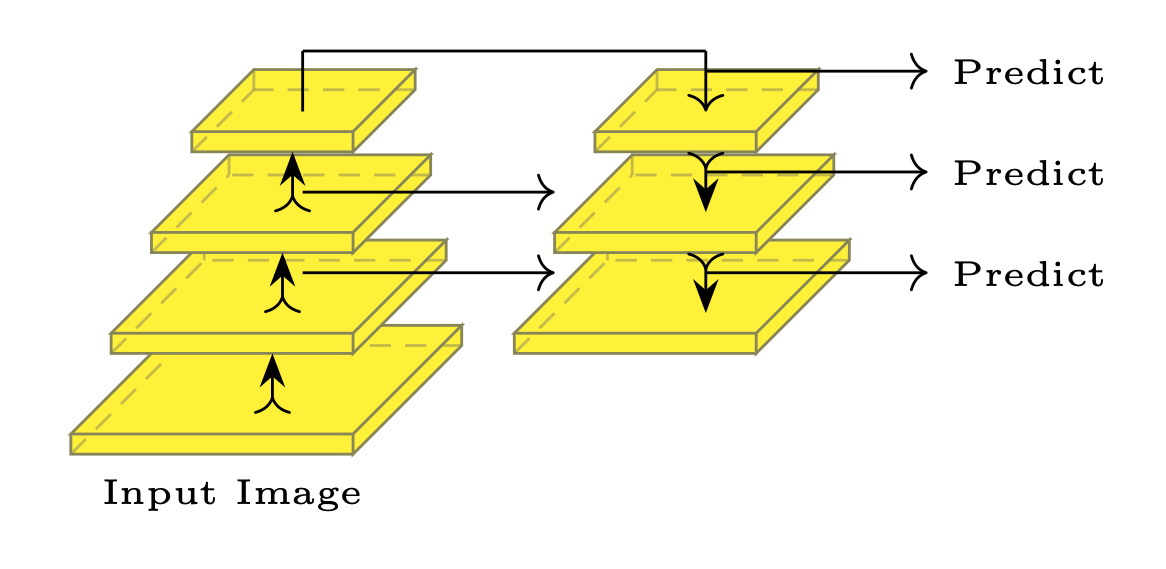}
\caption{}
\label{fig:FPN}
\end{subfigure}
\begin{subfigure}[b]{0.55\textwidth}
\centering
\includegraphics[width=\textwidth]{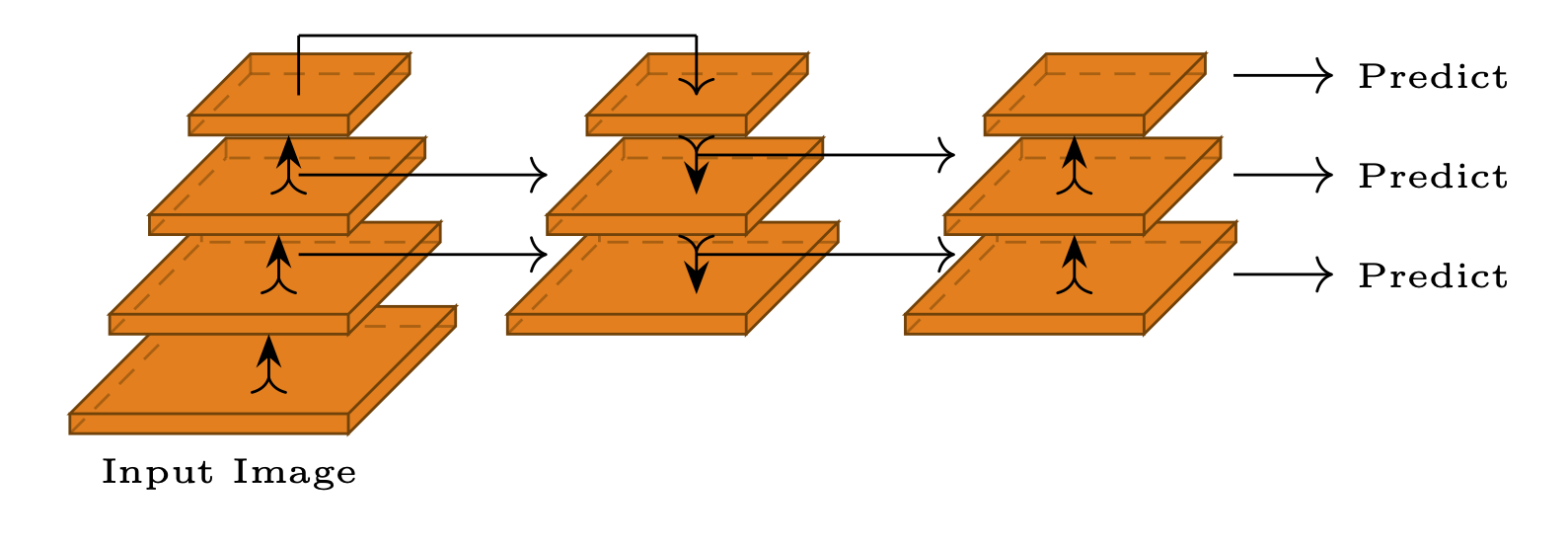}
\caption{}
\label{fig:PANet}
\end{subfigure}
\label{fig:FPN-PANet}
\caption{Schematic representations of $\colon$\textbf{(a)} Feature Pyramid Network (FPN)~\cite{LDGHHB2016,CMHM2022}, showing the convolution layers starting from the image layer located side by side to the constructed pyramid. The up-sampling process of the pyramid starts from the most top feature maps of the network containing more semantic contents. Each up-sample level of the pyramid is concatenated with its corresponding convoluted feature maps from the network. \textbf{(b)} Path Aggregation Network (PANet)~\cite{LQQSJ2018} comprised of the network itself along with two constructed pyramids. The first pyramid located next to the network is established similar to FPN that itself contributes to the creation of the new second bottom-up pyramid. Once again, across the second pyramid, concatenations are performed among the down-sampled levels with their corresponding feature maps from the first pyramid.}
\end{figure}

In contrast to YOLOv3, an $80\%$ reduction of the computational bottleneck was reported in~\cite{WLYWCH2019} by replacing Darknet-$53$~\cite{darknet} with Cross Stage Partial Networks (CSPNet), concerning COCO dataset~\cite{LMBBGHPRDZ2014}. The CSPNet was adopted as the backbone of the YOLOv5 along with the same FPN strategy from YOLOv3 to construct a sequence of two pyramids each comprised of four number of scaled levels. The most top level of the first pyramid is adopted from the convoluted network feature maps with the lowest resolution from CSPNet that are enriched with semantic features. The aim is to move from the chosen most top pyramid level down and establish the entire pyramid. This is done by adopting an up-sampling process with factor of $2$, while the up-sampled levels are concatenated with their corresponding layer from the network, in terms of spatial dimensions. In total, a pyramid of four levels are produced such that the lowest level has the highest dimension. The second pyramid is constructed similar to the first one, except this time the most bottom level of the first pyramid is chosen to start with and considered also as the most bottom level of the second pyramid. Next, three down-sampling process each by factor of $2$ are applied consequently to produce the next three levels of the second pyramid. Meanwhile, each produced level of the second pyramid gets concatenated with its corresponding level from the first pyramid, again in terms of spatial dimensions. This process of pyramid construction is referred in~\cite{YOLOv4} as ``Path Aggregation Network'' (PANet)~\cite{LQQSJ2018} (See Fig.~\ref{fig:PANet}) and also used in YOLOv5. As reported in~\cite{LQQSJ2018}, $256$ number of feature maps are consistently used during the pyramid constructions. A closer look at FPN and PANet reveals that both use lateral connections to boost information flow from bottom layers to top layers of their corresponding convolution-based networks namely Darknet-$53$~\cite{darknet} and CSPNet~\cite{WLYWCH2019} respectively, with a major difference that the latter lateral connections pass through only $8$ number of layers constructing the pyramid pairs but the former one passes through more than $100$ number of layers comprising the Darknet-$53$~\cite{darknet}. 
To perform dense prediction at the last YOLOv5 tensor layer, the same anchor based tensor structure in YOLOv3 is adopted.
\section{Detection Results}
To this end, we reveal our detection results obtained by varying both the YOLO architecture and also augmentation techniques~\cite{AetAl2020}. The model variation implemented by adopting four different architectures, namely YOLOv3 and YOLOv5 in small and larges sizes, that we abbreviate as YOLOv3s, YOLOv3l, YOLOv5s and YOLOv5l, respectively. All four models are trained using~\cite{ultralytics} with an early stopping strategy in case the mAP measure does not change after $100$ epochs. Each model is trained, validated and tested on $80\%$, $10\%$ and $10\%$ partitions of all images, to be the percentages of training, validation and testing sets with respect to the entire data set.

Before we start a detailed discussion about our obtained classification results, we would like to once again assert a high degree of difficulty that we face in this task that lies on the basis of a very strong correlation that exists among the colors of our work pieces in contrast to their background, namely the color of the factory. Such a similarity lead us to design two rigorous training and validation strategies$\colon$
\begin{enumerate}[(i)]
    \item First we train and validate on the entire $4900$ number of images comprised of original and augmented data and then we test on specifically $700$ augmented data set with a particular technique merged with its corresponding original set that itself contains also $700$ images. 
    \item Second, we target a particular augmentation technique and merge its augmented results with their original versions, in total $1400$ number of images to perform train, validation and test phases.
\end{enumerate}
The motivation behind the first case is to see which architecture is more robust against the environmental variations that introduced by augmented set of images used in test phase as unseen data. The second approach will provide us with the insights on which augmentation technique suits which architecture and better compensate for the color correlation issue that we face within our images.

With these strategies, we managed to report optimally trained models towards making correct 
classification with a high range of classification rates measured based on~\cite{EEGWWZ2015} and~\cite{LMBBGHPRDZ2014}. In current study we used two accuracy measures mAP$@0.5$~\cite{EEGWWZ2015} and mAP$@\left[0.5:0.95\right]$~\cite{LMBBGHPRDZ2014} introduced by PASCAL VOC and COCO challenges. As a general observation, our results are always better, if we consider a threshold of $0.5$ concerning the amount of overlap between IOUs, compared to an evaluation with a ten step varying threshold in range of $\left[0.5,0.95\right]$ with steps of $0.05$. However, because of the latter measure to be more robust and informative, we stay with it while reporting our best results in coming paragraphs.

\begin{figure} 
\centering
\begin{minipage}{.44\textwidth}
\subfloat[][YOLOv3s]
{
\includegraphics[width=.8\textwidth]{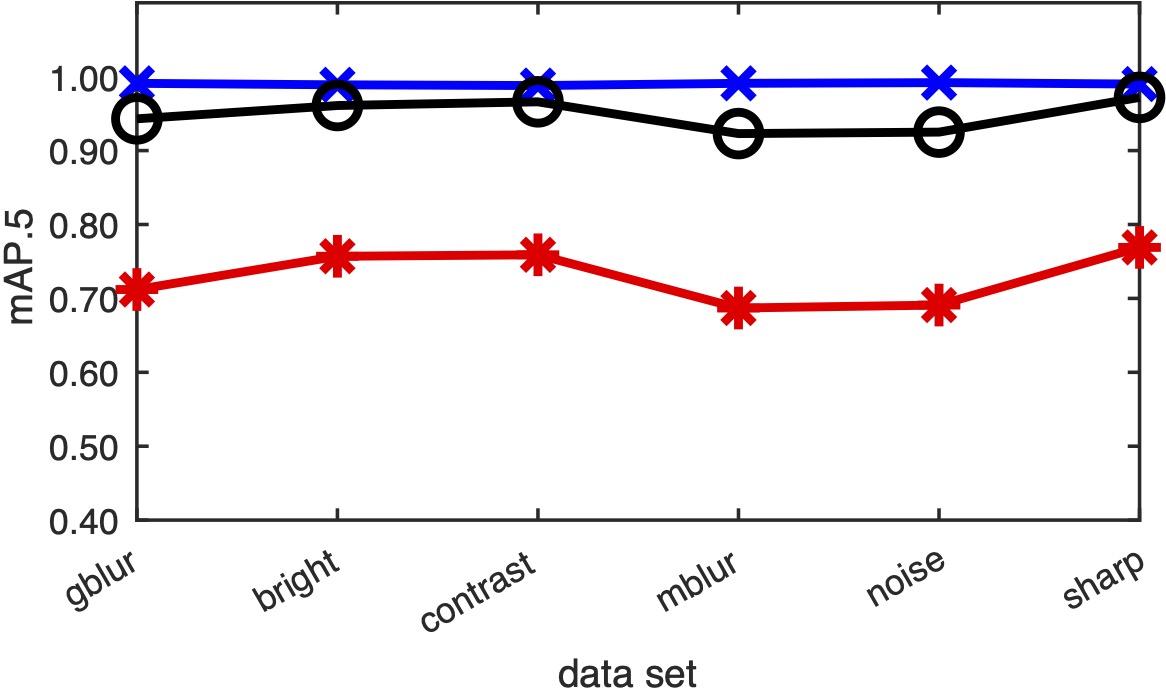}
\label{fig:PiecesYOLOv3A}
}
\\
\subfloat[][YOLOv3l]
{
\includegraphics[width=.8\textwidth]{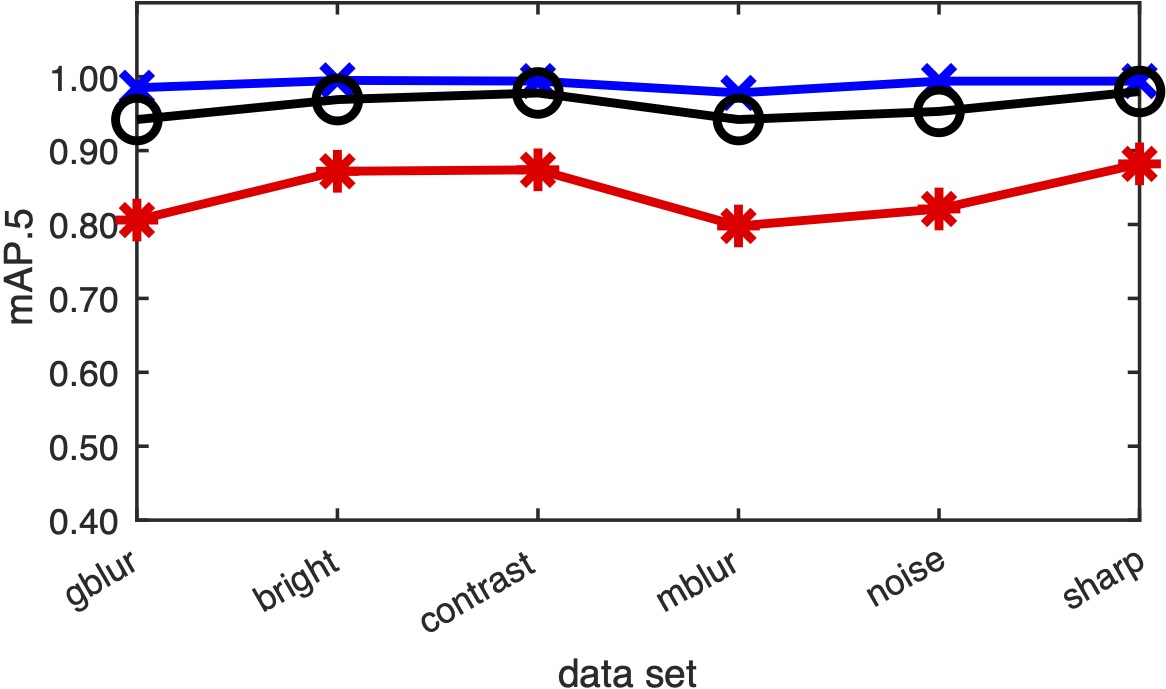}
\label{fig:PiecesYOLOv3B}
}
\end{minipage}
\qquad
\begin{minipage}{0.44\textwidth}
\subfloat[][YOLOv3s]
{
\includegraphics[width=.8\textwidth]{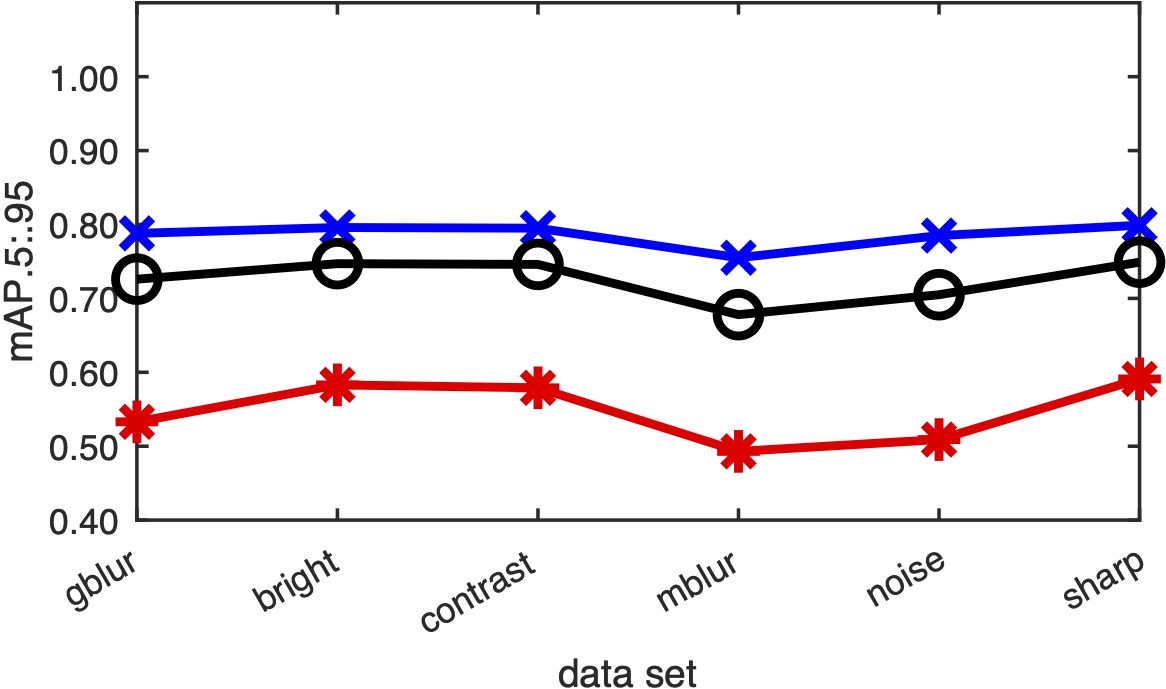}
\label{fig:PiecesYOLOv3C}
} 
\\
\subfloat[][YOLOv3l]
{
\includegraphics[width=.8\textwidth]{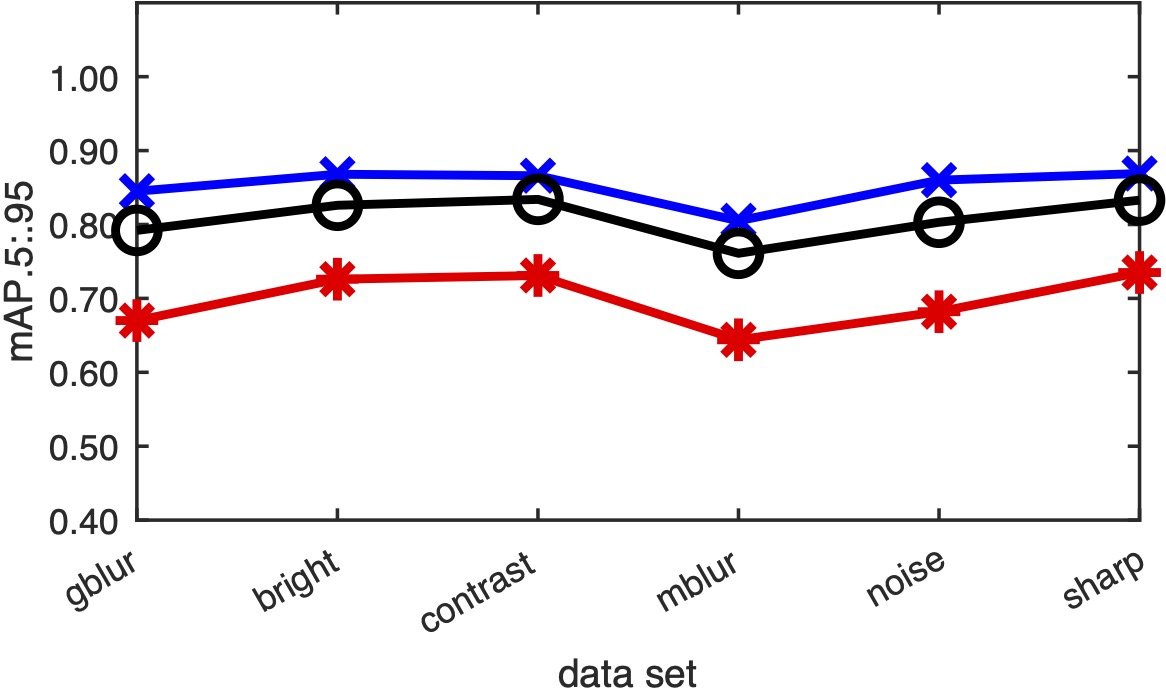}
\label{fig:PiecesYOLOv3D}
}
\end{minipage}
\caption{The classification rates mAP$@0.5$ shown as (a) and (b) versus mAP$@\left[0.5:0.95\right]$ shown as (c) and (d) revealing the efficiency of the trained YOLOv3s and YOLOv3l models in classifying the red, blue and white pieces shown as red, blue and black curves, respectively. Choosing the latter measure to be more informative, YOLO3vl shows the highest rate compared to (c) and also those plots shown as Figs.~\ref{fig:PiecesYOLOv5C} and~\ref{fig:PiecesYOLOv5D} corresponding to YOLOv5s and YOLOv5l when it comes to three particular augmentation techniques$\colon$ brightness, linear contrast and sharpness enhancements. Note that while varying the augmentation techniques we used strategy (ii) to implement the train, validate and test phases.}
\label{fig:PiecesYOLOv3}
\end{figure}

\begin{figure} 
\centering
\begin{minipage}{.44\textwidth}
\subfloat[][YOLOv5s]
{
\includegraphics[width=.8\textwidth]{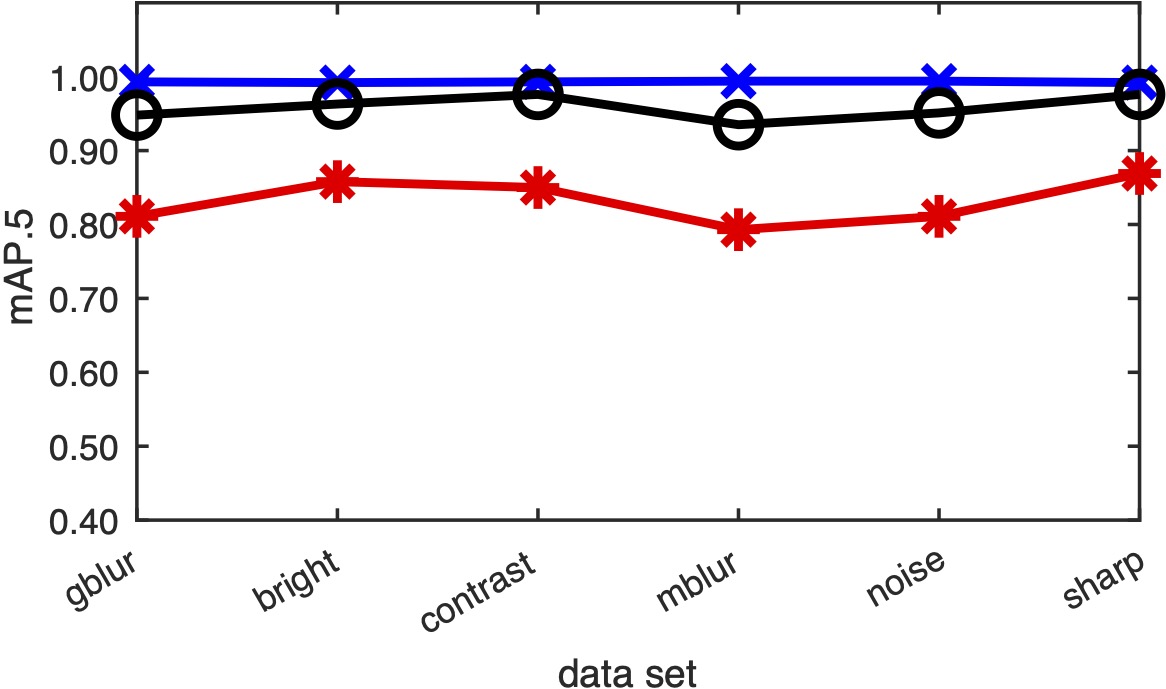}
\label{fig:PiecesYOLOv5A}
}
\\
\subfloat[][YOLOv5l]
{
\includegraphics[width=.8\textwidth]{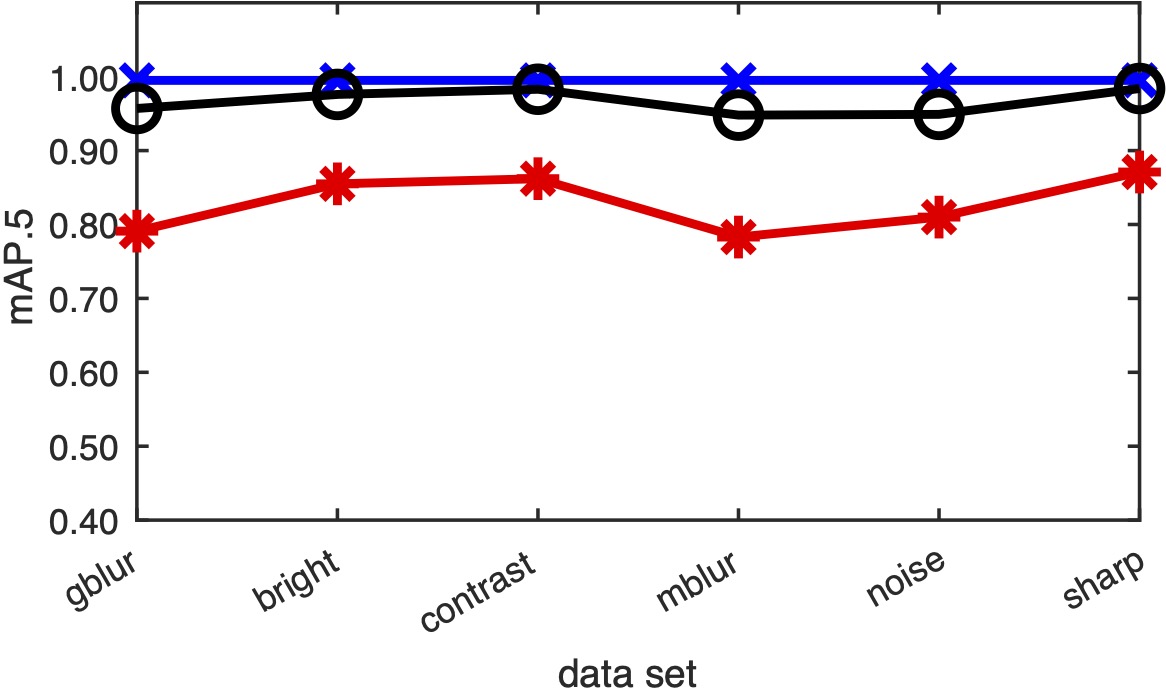}
\label{fig:PiecesYOLOv5B}
}
\end{minipage}
\qquad
\begin{minipage}{0.44\textwidth}
\subfloat[][YOLOv5s]
{
\includegraphics[width=.8\textwidth]{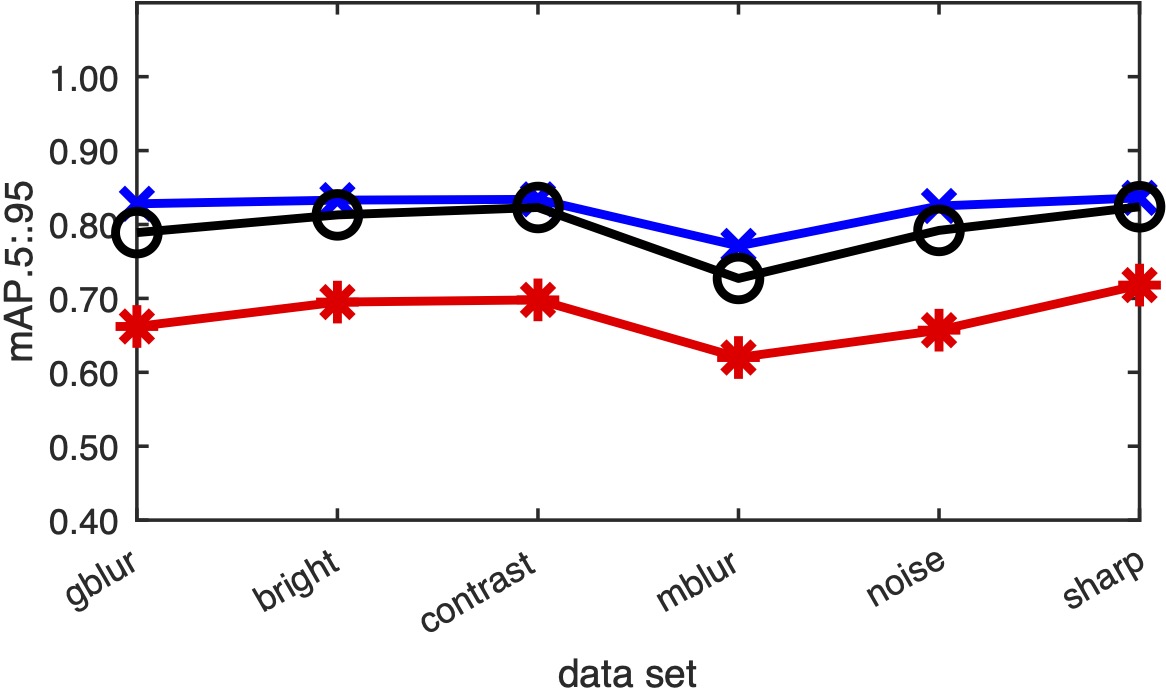}
\label{fig:PiecesYOLOv5C}
} 
\\
\subfloat[][YOLOv5l]
{
\includegraphics[width=.8\textwidth]{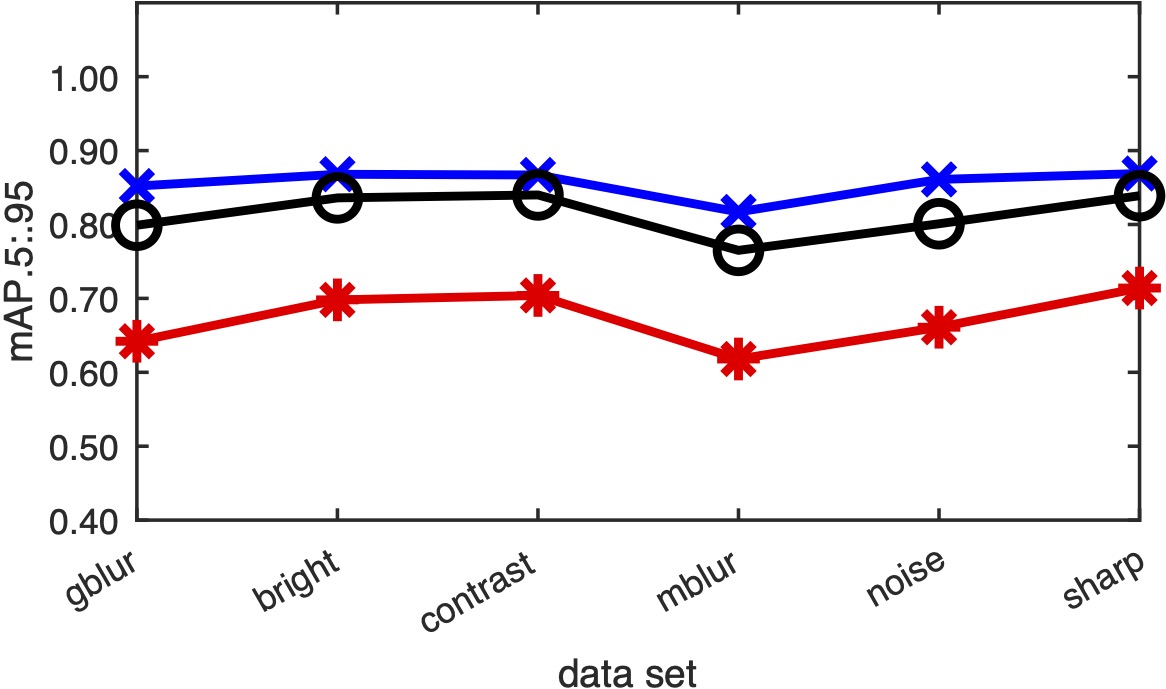}
\label{fig:PiecesYOLOv5D}
}
\end{minipage}
\caption{The classification rates mAP$@0.5$ shown as (a) and (b) versus mAP$@\left[0.5:0.95\right]$ shown as (c) and (d) revealing the efficiency of the trained YOLOv5s and YOLOv5l in classifying the red, blue and white pieces while different augmentation approaches are adopted based on (ii) strategy. We stay with the latter measure and observe that both YOLO5s and YOLOv5l detect the red and blue pieces shown as red and blue curves along with their corresponding measures. The white piece classification measures are also visualised as black curve showing acceptable ranges. However, by comparing both trained YOLO5 models in (c) and (d) with the one shown as Fig~\ref{fig:PiecesYOLOv3D} motivates us to select YOLOv3l to be the most practical model when it comes to specifically detecting the red pieces.}
\label{fig:PiecesYOLOv5}
\end{figure}

As the first concern, let us observe how the trained models are robust concerning the detection of the red, blue and white colors of the work pieces in contrast to their backgrounds as our model factory itself is mostly colored in blue and red colors. We chose to visualise the results corresponding to the blue and the red pieces as blue and red curves, respectively, and the white pieces as black curve as appears in Figs.~\ref{fig:PiecesYOLOv3} and~\ref{fig:PiecesYOLOv5}. Our results shown as the left columns of the Figs.~\ref{fig:PiecesYOLOv3} and~\ref{fig:PiecesYOLOv5} validate that all the models, except the YOLOv3s, have almost a mAP$@0.5$ of $0.80\%$ or higher while detecting varying piece colors. Here, in all cases the red work piece has always the lowest amount of mAP$@0.5$, though the blue and the white ones are detected with good values of the same measure. 

The best model based on mAP$@\left[0.5:0.95\right]$ is chosen to be YOLOv3l shown in Fig.~\ref{fig:PiecesYOLOv3D}, as the blue and the white exhibit a mean average precision almost equal or bigger then $80\%$, while the red pieces are managed to be detected with a rate of almost always in vicinity of $70\%$. In specific and within the Fig.~\ref{fig:PiecesYOLOv3D} the red curve representing the red pieces clearly is located above the $70\%$ when it comes to brightness, linear contrast and sharpness augmentations used to enhance the color effects of the images. The mAP$@\left[0.5:0.95\right]$ values corresponding to these three augmentation techniques in Figs.~\ref{fig:PiecesYOLOv5C} and~\ref{fig:PiecesYOLOv5D} are clearly almost equal or below $70\%$. This motivates us to chose the YOLOv3l to be more practical and robust compared to YOLOv5s and YOLOv5l, when it comes to detecting specifically the red and in general the blue and the white pieces. A very meaningful comparison comes to the picture where YOLOv3l shows its practicability in contrast to YOLOv5s and YOLOv5l concerning the sharpness augmentation technique. Here, the trained YOLOv3l model in Fig.~\ref{fig:PiecesYOLOv3D} shows a distinct sharpness measure above $70\%$, where as the same quantity is almost equal to $70\%$ corresponding to YOLOv5s and YOLOv5l shown as Figs.~\ref{fig:PiecesYOLOv5C} and~\ref{fig:PiecesYOLOv5D}. With this, we conclude the suitability of our training strategy (ii) to show the sharpness enhancement technique to be the most effective one in direction of obtaining the highest degree of robustness against the color correlation aspect of out data. Note that, the YOLO3s model shown as Fig.~\ref{fig:PiecesYOLOv3C} shows poor result in contrast to other adopted YOLO model in current study and concerning the mAP$@\left[0.5:0.95\right]$ measure. 
%
%

\begin{figure}
  \begin{minipage}[c]{0.6\textwidth}
    \includegraphics[width=.44\textwidth]{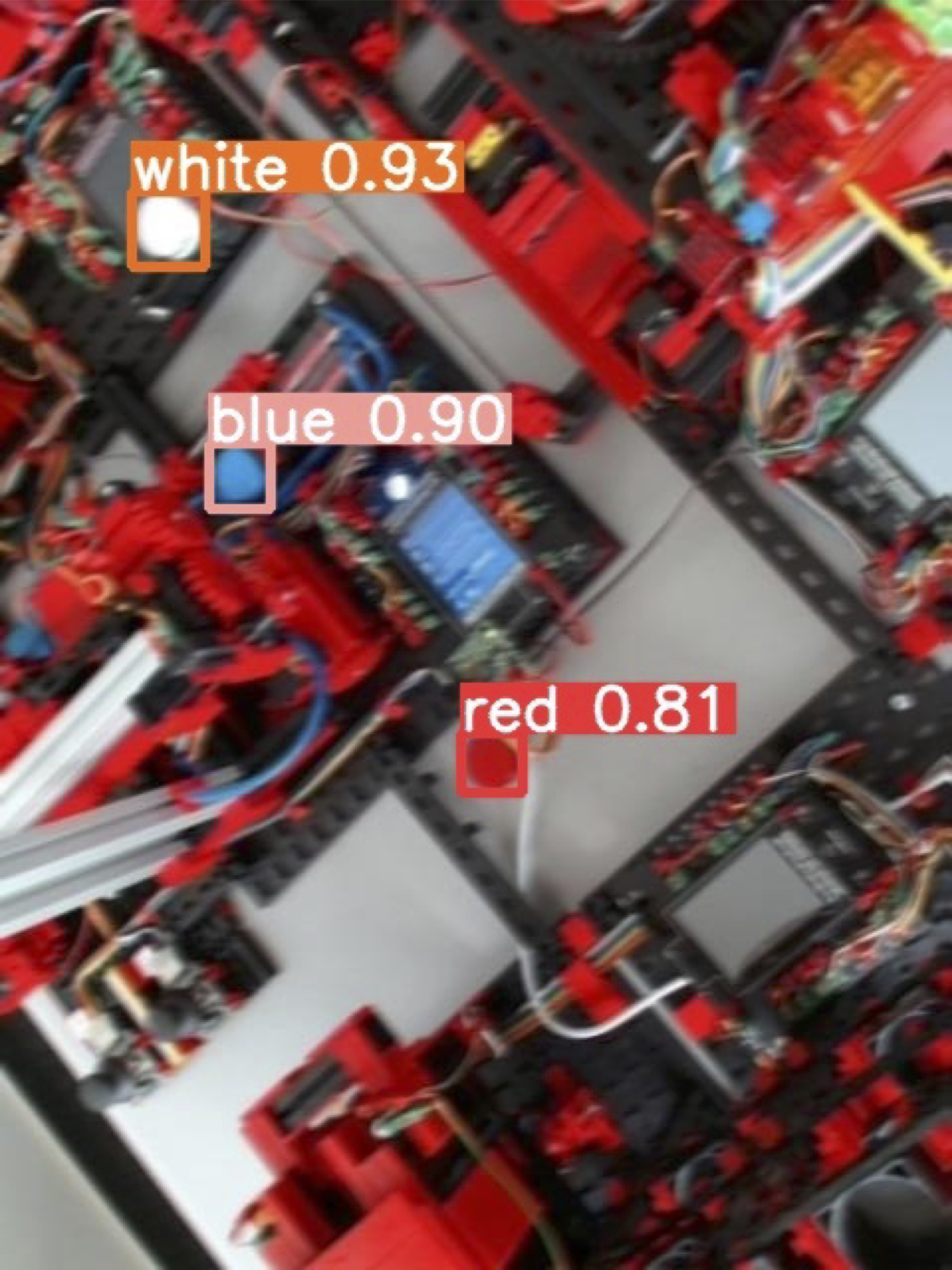}
    \includegraphics[width=.44\textwidth]{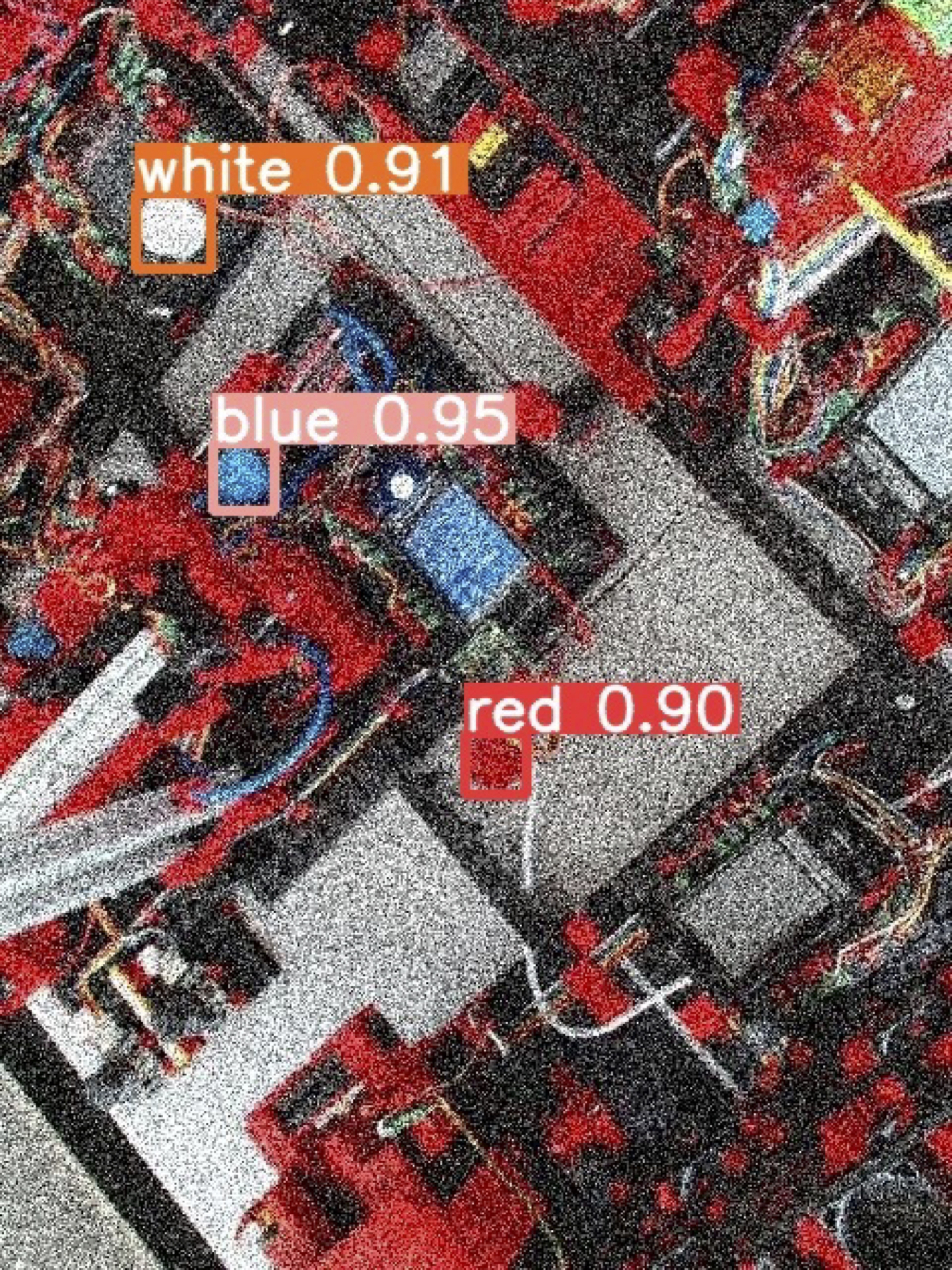}
  \end{minipage}\hfill
  \begin{minipage}[c]{0.4\textwidth}
\caption{A pair of images along with three color work pieces undergone through motion blur and noise perturbation while successfully being classified by YOLOv5s, as we expect this architecture to have the highest classification rate according to the light blue curve produced based on strategy (ii) shown as Fig.~\ref{fig:DataYOLOv5C}.}
\label{fig:YOLOv5sResults}
  \end{minipage}
\end{figure}

\begin{figure}
  \begin{minipage}[c]{0.6\textwidth}
    \includegraphics[width=.48\textwidth]{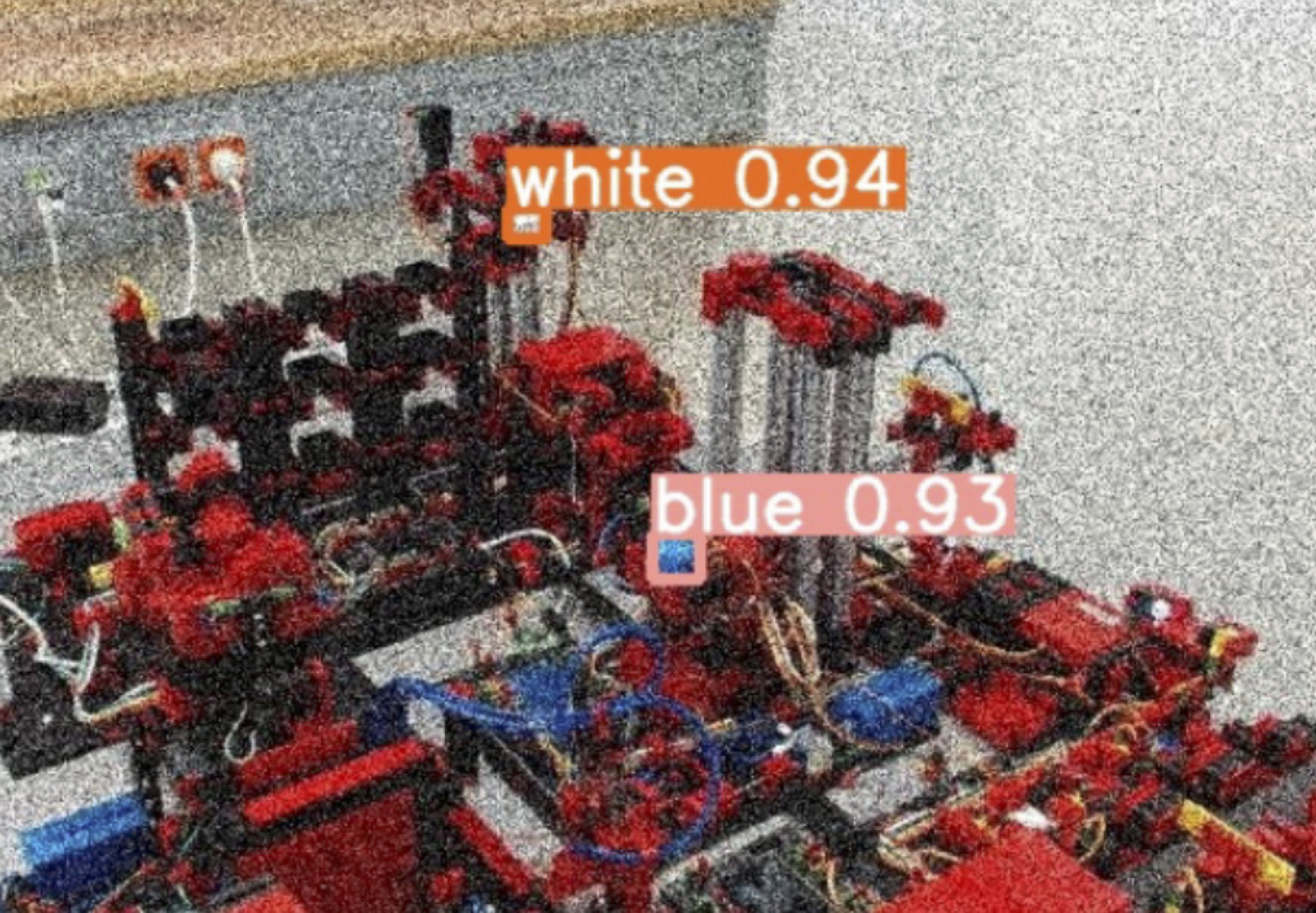}
    \includegraphics[width=.44\textwidth]{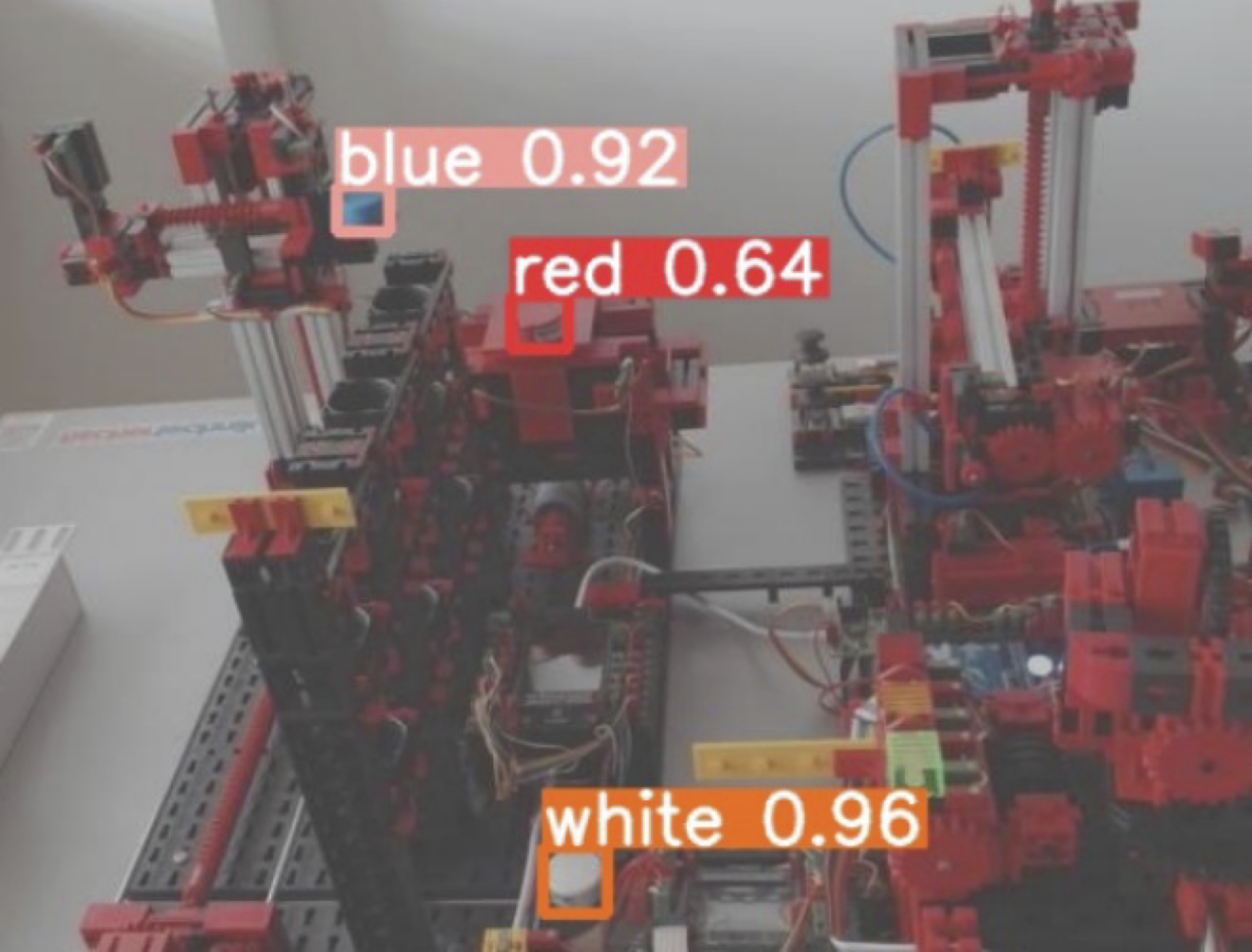}
  \end{minipage}\hfill
  \begin{minipage}[c]{0.4\textwidth}
\caption{Two augmented images using (a) noise perturbation and (b) linear contrast adjustments containing color work pieces that are successfully detected by YOLOv3l model trained based on strategy (i).}
\label{fig:YOLOv3Results}
  \end{minipage}
\end{figure}

\begin{figure}
  \begin{minipage}[c]{0.6\textwidth}
    \includegraphics[width=.44\textwidth]{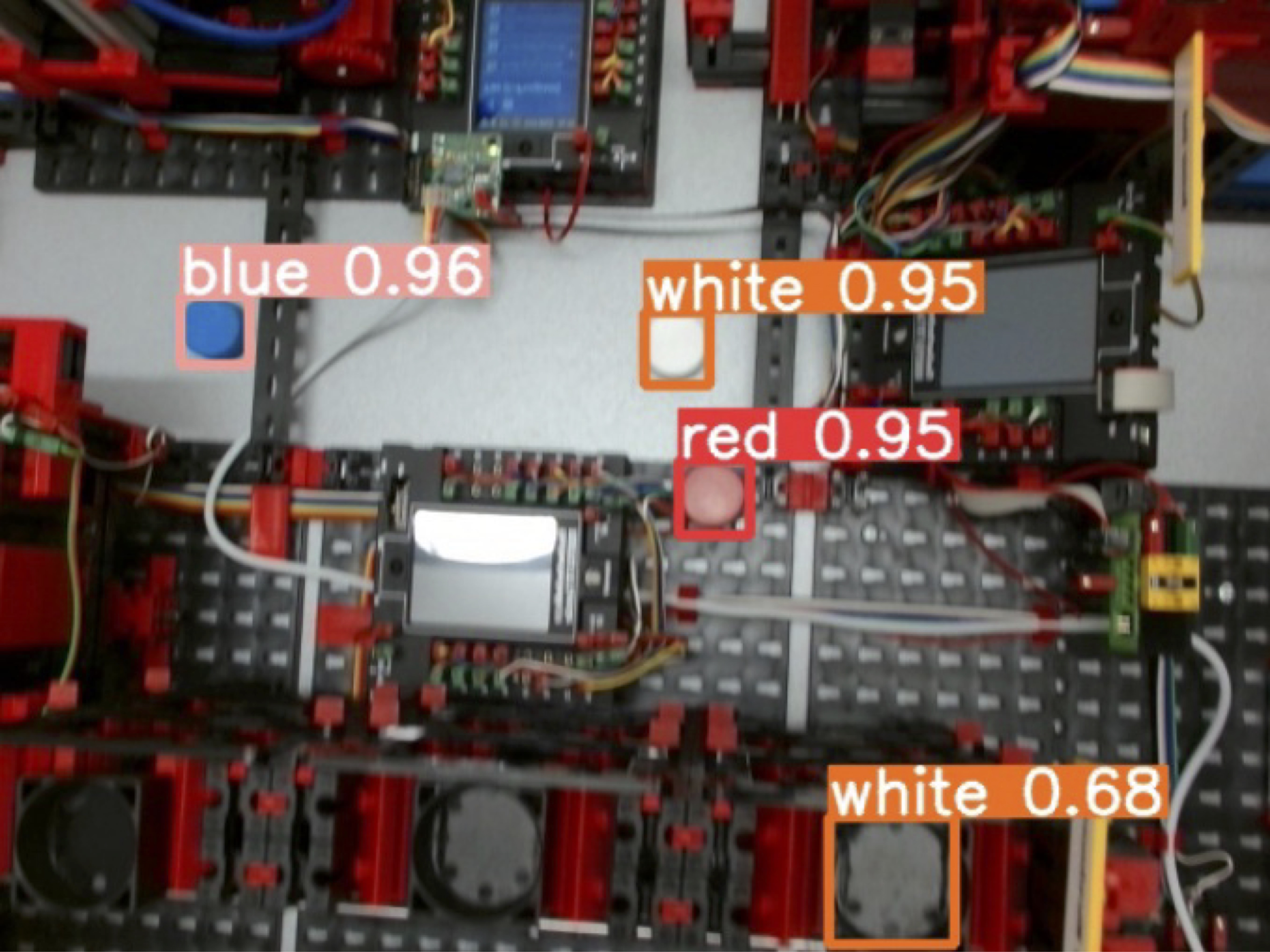}
    \includegraphics[width=.47\textwidth]{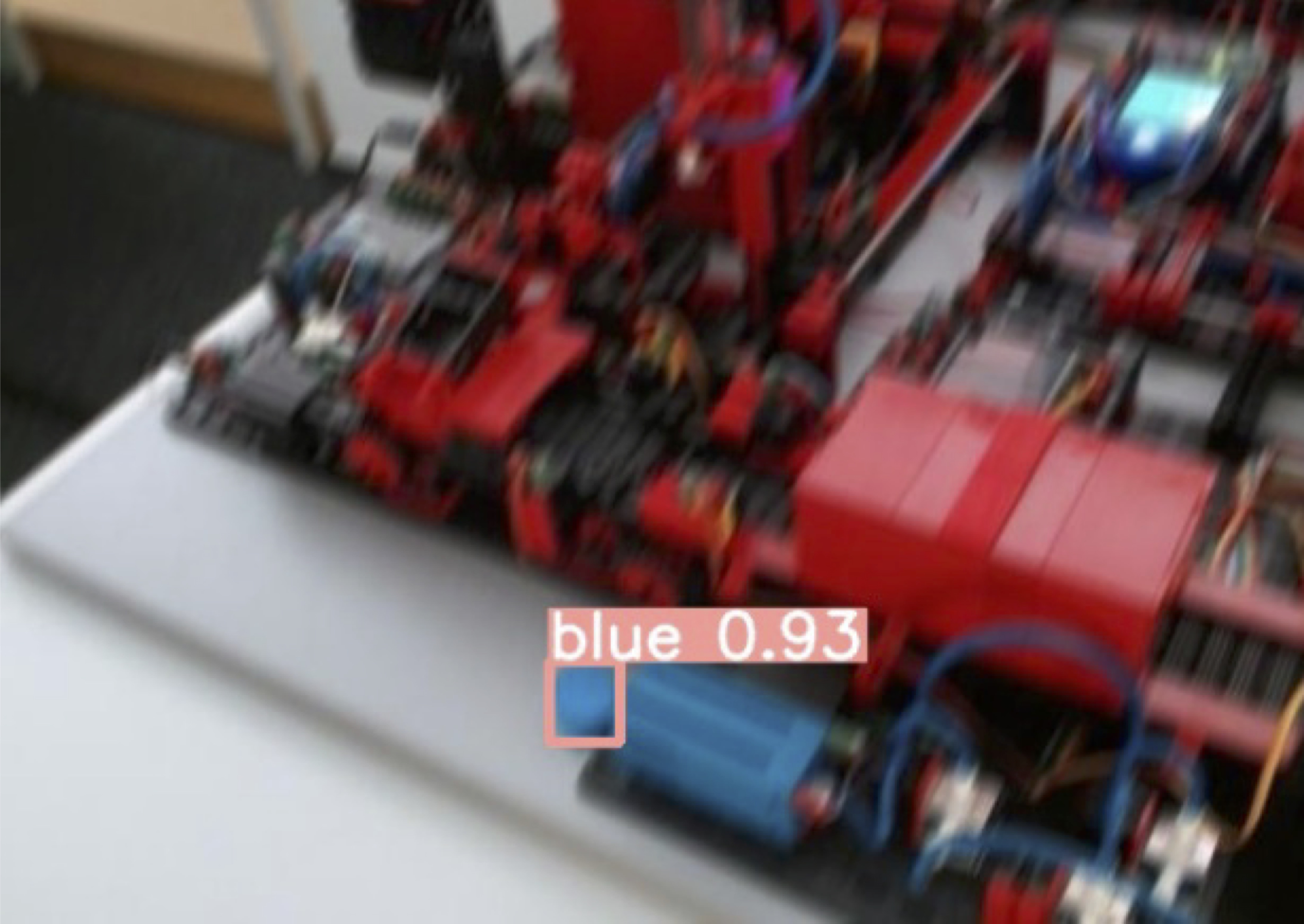}
  \end{minipage}\hfill
  \begin{minipage}[c]{0.4\textwidth}
\caption{Two augmented images using (a) Gaussian blur and (b) motion blur containing falsely classified work pieces as the result of applying YOLOv3l model trained based on strategy (i).}
\label{fig:FalseAcceptence}
  \end{minipage}
\end{figure}

\begin{figure} 
\centering
\begin{minipage}{.44\textwidth}
\subfloat[][YOLOv3s]
{
\includegraphics[width=.8\textwidth]{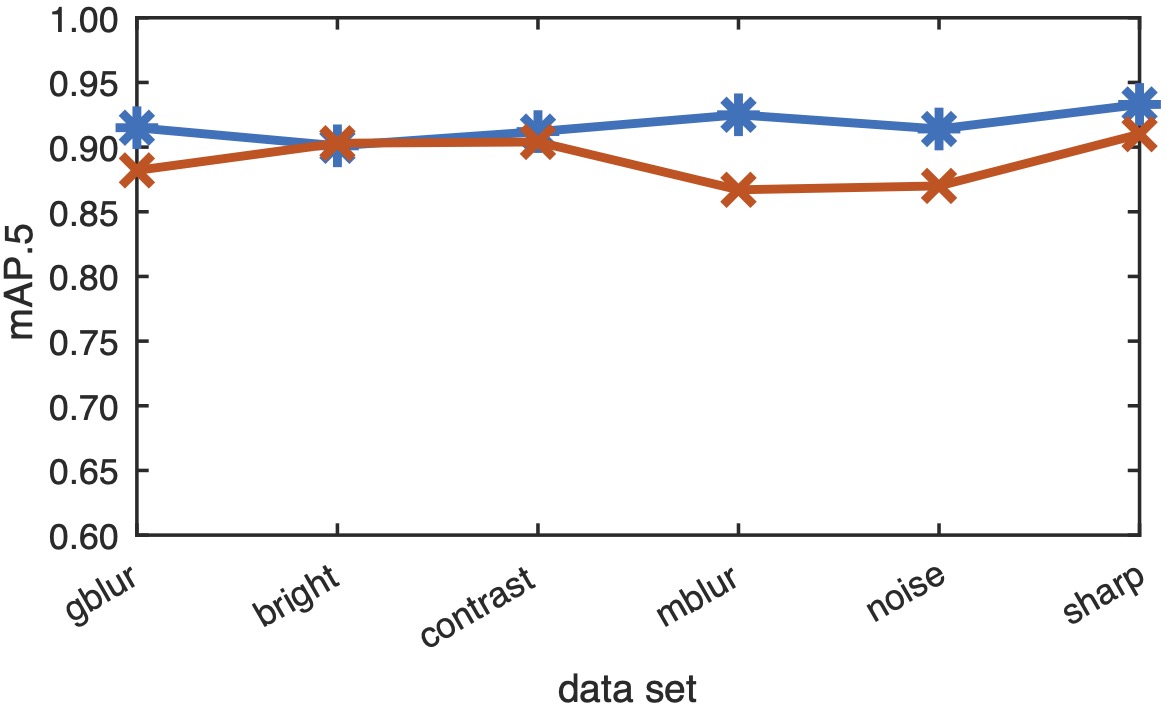}
\label{fig:DataYOLOv3A}
}
\\
\subfloat[][YOLOv3l]
{
\includegraphics[width=.8\textwidth]{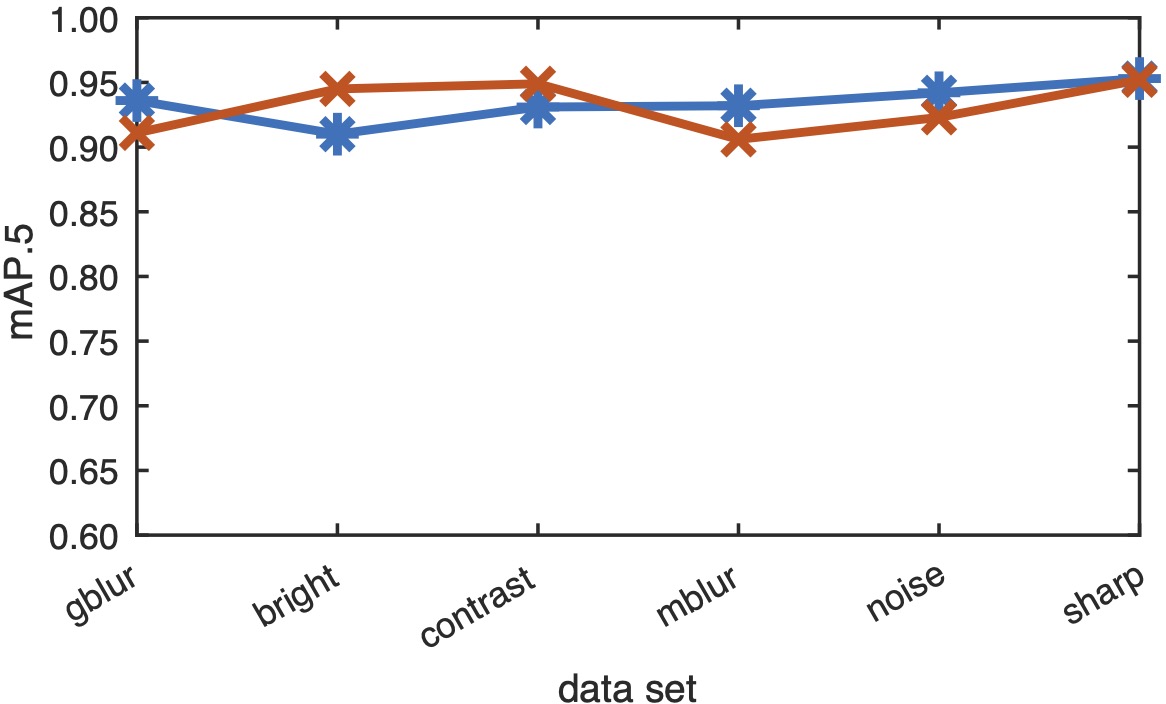}
\label{fig:DataYOLOv3B}
}
\end{minipage}
\qquad
\begin{minipage}{0.44\textwidth}
\subfloat[][YOLOv3s]
{
\includegraphics[width=.8\textwidth]{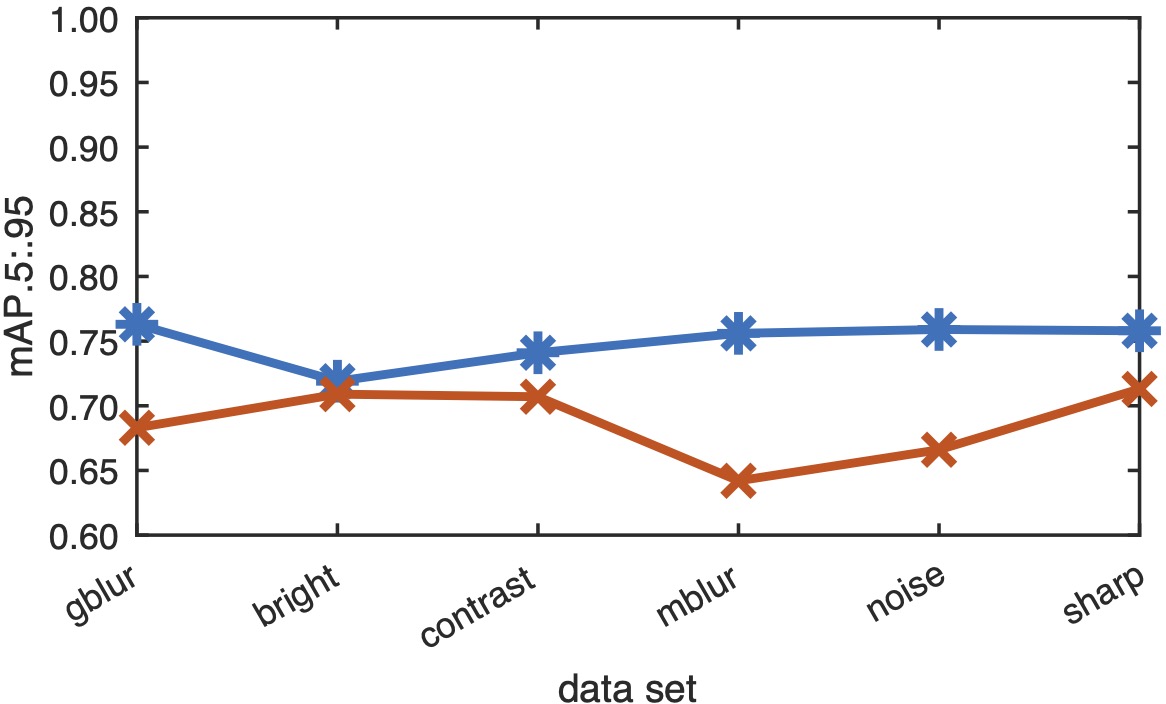}
\label{fig:DataYOLOv3C}
} 
\\
\subfloat[][YOLOv3l]
{
\includegraphics[width=.8\textwidth]{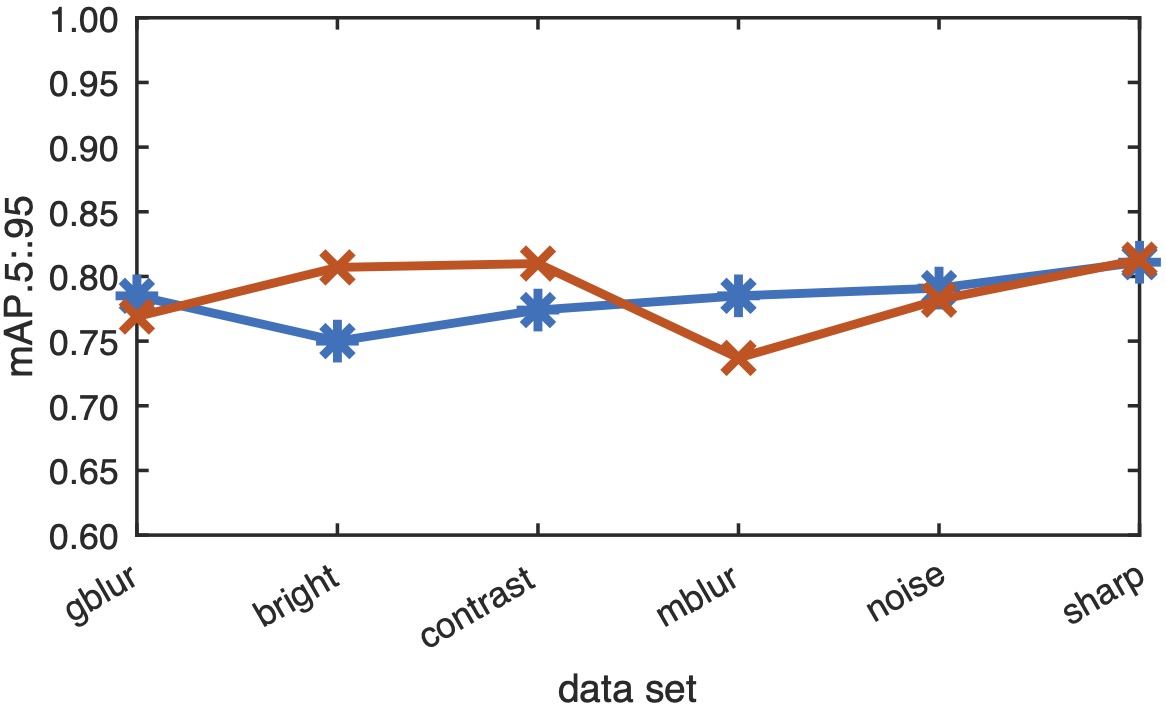}
\label{fig:DataYOLOv3D}
}
\end{minipage}
\caption{The classification results of YOLOv3s and YOLOv3l trained on both strategies (\romannum{1}) and (\romannum{2}) while considering mAP$@0.5$ (a) and (b) and mAP$@\left[0.5:0.95\right]$ (c) and (d). By staying with the second measure we conclude that by following the (i) strategy the YOLOv3l shown as orange curve in (d) has the highest rates compared to (c) and also plots shown as Figs.~\ref{fig:DataYOLOv5C} and~\ref{fig:DataYOLOv5D} corresponding to YOLOv5s and YOLOv5l in particular concerning the brightness, linear contrast and sharpness enhancement techniques.}\label{fig:DataYOLOv3}
\end{figure}

\begin{figure} 
\centering
\begin{minipage}{.44\textwidth}
\subfloat[][YOLOv5s]
{
\includegraphics[width=.8\textwidth]{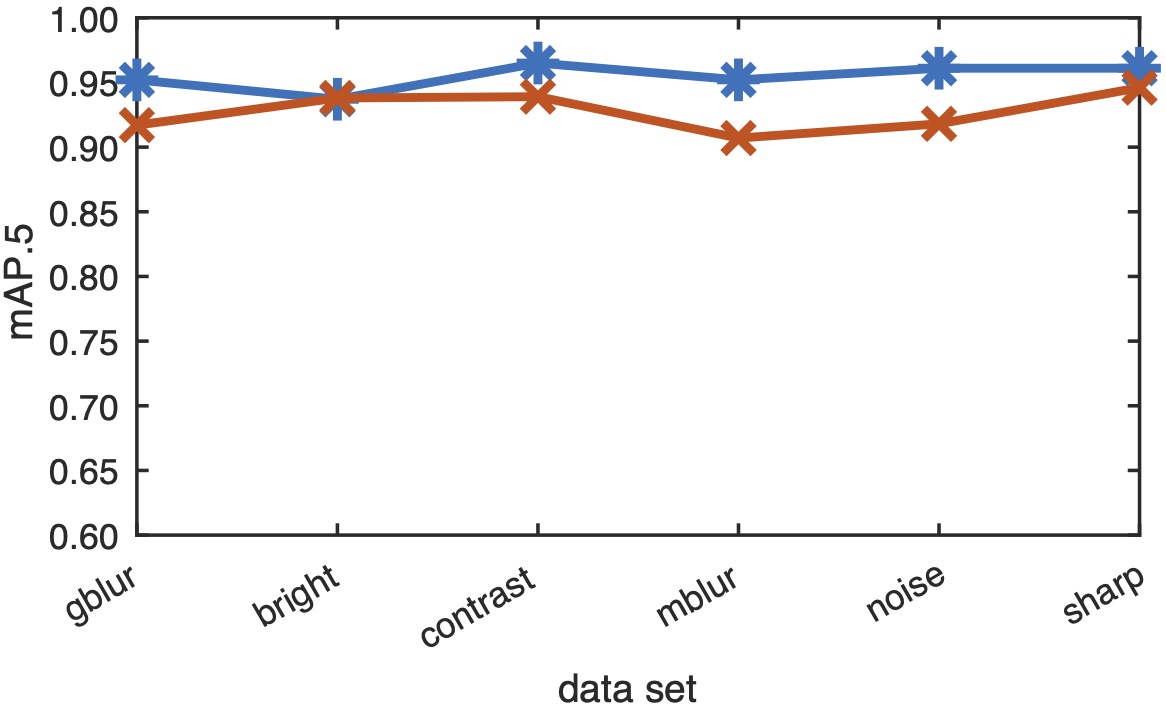}
\label{fig:DataYOLOv5A}
}
\\
\subfloat[][YOLOv5l]
{
\includegraphics[width=.8\textwidth]{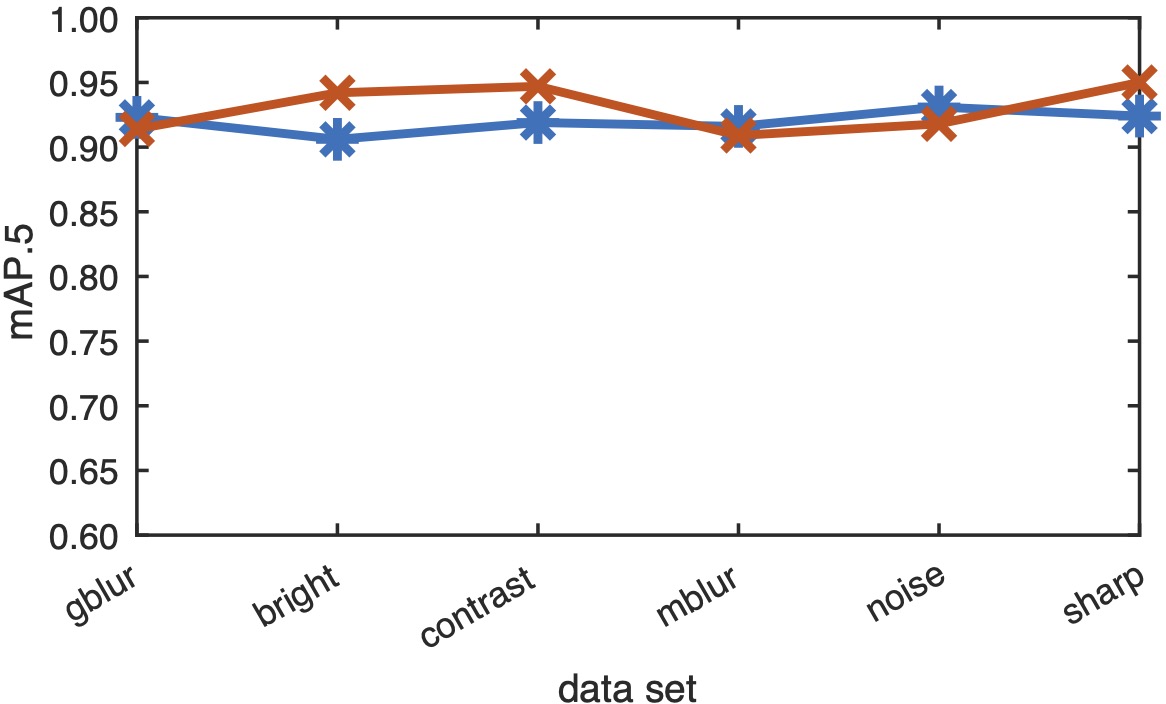}
\label{fig:DataYOLOv5B}
}
\end{minipage}
\qquad
\begin{minipage}{0.44\textwidth}
\subfloat[][YOLOv5s]
{
\includegraphics[width=.8\textwidth]{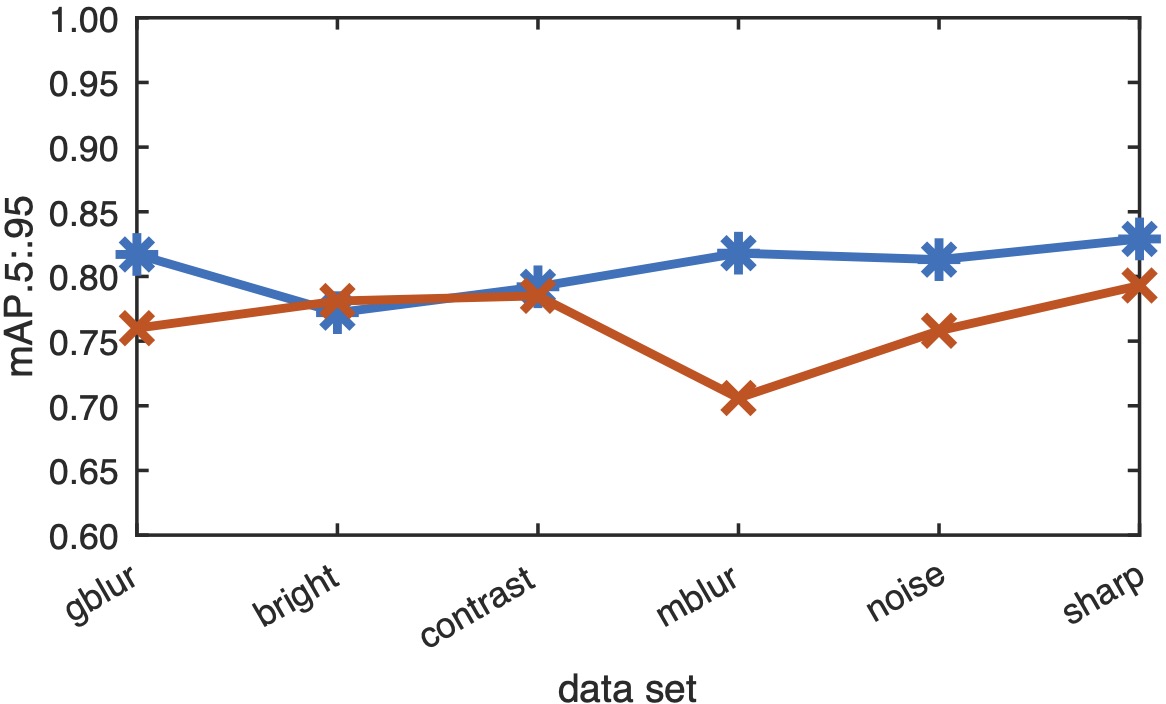}
\label{fig:DataYOLOv5C}
} 
\\
\subfloat[][YOLOv5l]
{
\includegraphics[width=.8\textwidth]{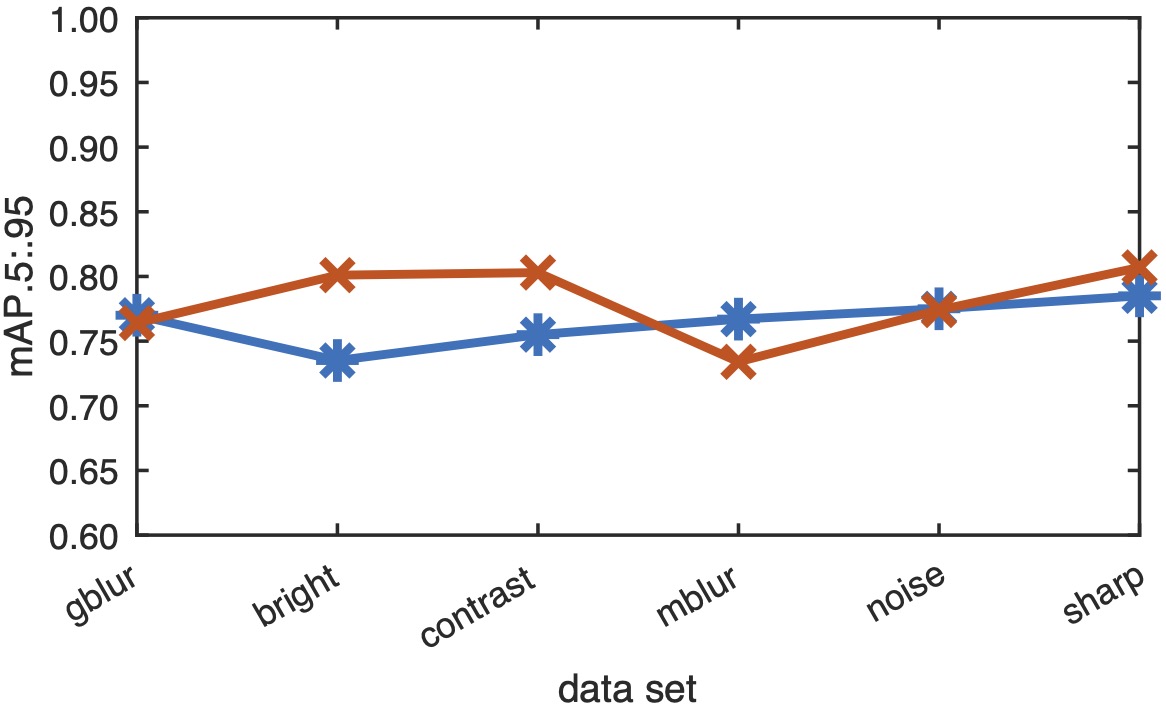}
\label{fig:DataYOLOv5D}
}
\end{minipage}
\caption{The classification results of YOLOv5s and YOLOv5l trained on both strategies (\romannum{1}) and (\romannum{2}) while considering mAP$@0.5$ (a) and (b) and mAP$@\left[0.5:0.95\right]$ (c) and (d). We stay with the second measure and conclude that by following the (ii) strategy the YOLOv5s shown as light blue curve in (c) has the highest rates compared to (d) and also plots shown as Figs.~\ref{fig:DataYOLOv3C} and~\ref{fig:DataYOLOv3D} corresponding to YOLOv3s and YOLOv3l concerning almost all augmentation techniques. However, training on the entire $4900$ images and validating and testing on particular augmented data set, namely following strategy (i), results to YOLOv5l to be the most efficient model shown as orange curve in (d), specifically concerning the brightness, linear contrast and sharpness enhancement operations.}
\label{fig:DataYOLOv5}
\end{figure}

Let us proceed with the second stage of our investigation and see the impact of the augmentation techniques realised on each of the model architectures, concerning the mAP$@\left[0.5:0.95\right]$ measure, as appeared in Figs.~\ref{fig:DataYOLOv3D},~\ref{fig:DataYOLOv5C} and~\ref{fig:DataYOLOv5D}. Here, we color code our training strategies (\romannum{1}) and (\romannum{2}) as orange and light blue. In general, the light blue curve shown in all plots of the Fig.~\ref{fig:DataYOLOv3} and~\ref{fig:DataYOLOv5} is established by training the corresponding model architecture trained, validated and tested on $1400$ number images, namely $700$ augmented with a particular technique along with their $700$ original counterparts. Our first observation is on Fig.~\ref{fig:DataYOLOv5C} shows that the light blue curve to maintain almost a classification rate higher than $80\%$, with respect to all augmentation techniques except the brightness enhancement operation. This means to tackle a particular scenario that can be simulated by an augmentation technique, namely a motion blur due to the movement of the camera across the factory field, a YOLOv5s trained with suitable motion blur augmentation technique will result to the best model. A pair of images undergone through motion blur and noise perturbation are shown as Fig.~\ref{fig:YOLOv5sResults} that are successfully classified by YOLOv5s. 
In contrast, the orange curves shown in Figs.~\ref{fig:DataYOLOv3D} and~\ref{fig:DataYOLOv5D} reveal us that, the bigger architectures YOLOv3l and YOLOv5l trained and validated on the entire set of $4900$ images and targeted to test each of the augmented sets are only efficient when it comes to three particular augmentation techniques, namely brightness, linear contrast and sharpness enhancement as their corresponding values are among the highest ans almost near to $80\%$ shown as Figs.~\ref{fig:DataYOLOv3D},~\ref{fig:DataYOLOv5C} and~\ref{fig:DataYOLOv5D}. This conclusion can be fully understood as the latter three techniques will result in more enhancement between the color similarity of the pieces and the factory, concerning the blue and the red colors. A pair of images highly adjusted by noise and linear contrast enhancement are shown to be successfully processed by YOLOv3l model along with their classified work pieces shown as Fig.~\ref{fig:YOLOv3Results}. Finally, let us reveal two falsely classified cases by YOLOv5l and YOLOv3s as Fig.~\ref{fig:FalseAcceptence}.
%
%
%
\section{Summary and Conclusions}
In this paper, we proposed an approach to detect and classify the work pieces transported across a Fischertechnik factory model based on their colors. Our approach benefits from the adoption of the YOLO models trained on an augmented set of images captured from the factory. As we perform a supervised classification, we label the ground truth work pieces, which means we captured the factory while operating on the work pieces and later labeled the captured images. To model the real world environments we used a set of augmentation techniques to account for sever distortion that may happen in real world.

We also conjectured that different YOLO architectures to behave differently on our augmented data, hence we opt to train four different architectures with two different training and validation strategies. With this, we aim to see if any training strategy along with a particular model may result in better classification results as our prepared data is highly correlated concerning the red and the blue work pieces since their colors are alike with the factory colors to be considered as the background. Hence and to be realistic, we expected low classification results concerning specifically the red work pieces that we also observed across our results.

In brief, two particular model YOLOv3 and YOLOv5 in large and small scales, respectively, shown to be optimum for particular purposes. The latter model, can be trained on small data sets and used on specific scenarios with highly severed environmental distortions while the former model shows good classification results concerning the cases that work pieces have color similarities with their backgrounds. 

In the future, a comprehensive study needs to be conducted to explore more insights on the cases where the YOLO models did not trained optimally to perform the detection and classification tasks. This can be carried on based on suitable pre-processing techniques to decompose the data images in uncorrelated color spaces and the impact that it may have on final object classification results.
\bibliography{ref}

\begin{thebibliography}{10}
\providecommand{\url}[1]{{#1}}
\providecommand{\urlprefix}{URL }
\expandafter\ifx\csname urlstyle\endcsname\relax
  \providecommand{\doi}[1]{DOI~\discretionary{}{}{}#1}\else
  \providecommand{\doi}{DOI~\discretionary{}{}{}\begingroup
  \urlstyle{rm}\Url}\fi

\bibitem{ultralytics}
Ultralytics github.
\newblock \url{https://github.com/ultralytics}.
\newblock Accessed: October 27, 2022

\bibitem{ACJetAl2017}
Abele, E., Chryssolouris, G., Sihn, W., Metternich, J., ElMaraghy, H.A.,
  Seliger, G., Sivard, G., Elmaraghy, W., Hummel, V., Tisch, M., Seifermann,
  S.: Learning factories for future oriented research and education in
  manufacturing.
\newblock Cirp Annals-manufacturing Technology \textbf{66}, 803--826 (2017)

\bibitem{BTAAK2018}
Benjdira, B., Khursheed, T., Koubaa, A., Ammar, A., Ouni, K.: Car detection
  using unmanned aerial vehicles: Comparison between faster r-cnn and yolov3
  (2018).
\newblock \doi{10.48550/ARXIV.1812.10968}.
\newblock \urlprefix\url{https://arxiv.org/abs/1812.10968}

\bibitem{darknet}
Bochkovskiy, A.:  \urlprefix\url{https://github.com/AlexeyAB/darknet}

\bibitem{YOLOv4}
Bochkovskiy, A., Wang, C., Liao, H.M.: Yolov4: Optimal speed and accuracy of
  object detection.
\newblock CoRR \textbf{abs/2004.10934} (2020).
\newblock \urlprefix\url{https://arxiv.org/abs/2004.10934}

\bibitem{CV1995}
Cortes, C., Vapnik, V.: Support-vector networks.
\newblock Machine learning \textbf{20}(3), 273--297 (1995)

\bibitem{ImageNet}
Deng, J., Dong, W., Socher, R., Li, L.J., Li, K., Fei-Fei, L.: Imagenet: A
  large-scale hierarchical image database.
\newblock In: 2009 IEEE conference on computer vision and pattern recognition,
  pp. 248--255. Ieee (2009)

\bibitem{EEGWWZ2015}
Everingham, M., Eslami, S.M., Gool, L., Williams, C.K., Winn, J., Zisserman,
  A.: The pascal visual object classes challenge: A retrospective.
\newblock Int. J. Comput. Vision \textbf{111}(1), 98–136 (2015).
\newblock \doi{10.1007/s11263-014-0733-5}.
\newblock \urlprefix\url{https://doi.org/10.1007/s11263-014-0733-5}

\bibitem{GDDM2013}
Girshick, R., Donahue, J., Darrell, T., Malik, J.: Rich feature hierarchies for
  accurate object detection and semantic segmentation (2013).
\newblock \doi{10.48550/ARXIV.1311.2524}.
\newblock \urlprefix\url{https://arxiv.org/abs/1311.2524}

\bibitem{JLetAl2022}
Grüger, J., Malburg, L., Mangler, J., Bertrand, Y., Rinderle-Ma, S., Bergmann,
  R., Asensio, E.S.: Sensorstream: An xes extension for enriching event logs
  with iot-sensor data (2022).
\newblock \doi{10.48550/ARXIV.2206.11392}.
\newblock \urlprefix\url{https://arxiv.org/abs/2206.11392}

\bibitem{CLAM2009}
Gu, C., Lim, J.J., Arbelaez, P., Malik, J.: Recognition using regions.
\newblock In: 2009 IEEE Conference on Computer Vision and Pattern Recognition,
  pp. 1030--1037 (2009).
\newblock \doi{10.1109/CVPR.2009.5206727}

\bibitem{HGDG2017}
He, K., Gkioxari, G., Dollár, P., Girshick, R.: Mask r-cnn (2017).
\newblock \doi{10.48550/ARXIV.1703.06870}.
\newblock \urlprefix\url{https://arxiv.org/abs/1703.06870}

\bibitem{YOLOv5}
Jocher, G., Chaurasia, A., Stoken, A., Borovec, J., NanoCode012, Kwon, Y.,
  TaoXie, Michael, K., Fang, J., imyhxy, Lorna, Wong, C., Yifu, Z., V, A.,
  Montes, D., Wang, Z., Fati, C., Nadar, J., Laughing, UnglvKitDe, tkianai,
  yxNONG, Skalski, P., Hogan, A., Strobel, M., Jain, M., Mammana, L., xylieong:
  {ultralytics/yolov5: v6.2 - YOLOv5 Classification Models, Apple M1,
  Reproducibility, ClearML and Deci.ai integrations} (2022).
\newblock \doi{10.5281/zenodo.7002879}.
\newblock \urlprefix\url{https://doi.org/10.5281/zenodo.7002879}

\bibitem{AetAl2020}
Jung, A.B., Wada, K., Crall, J., Tanaka, S., Graving, J., Reinders, C., Yadav,
  S., Banerjee, J., Vecsei, G., Kraft, A., Rui, Z., Borovec, J., Vallentin, C.,
  Zhydenko, S., Pfeiffer, K., Cook, B., Fernández, I., De~Rainville, F.M.,
  Weng, C.H., Ayala-Acevedo, A., Meudec, R., Laporte, M., et~al.: {Imgaug}.
\newblock \url{https://github.com/aleju/imgaug} (2020).
\newblock Online; accessed 01-Feb-2020

\bibitem{HG2022}
Jung, H.K., Choi, G.S.: Improved yolov5: Efficient object detection using drone
  images under various conditions.
\newblock Applied Sciences \textbf{12}(14) (2022).
\newblock \doi{10.3390/app12147255}.
\newblock \urlprefix\url{https://www.mdpi.com/2076-3417/12/14/7255}

\bibitem{JJS2020}
Kim, J.a., Sung, J.Y., Park, S.h.: Comparison of faster-rcnn, yolo, and ssd for
  real-time vehicle type recognition.
\newblock In: 2020 IEEE International Conference on Consumer Electronics - Asia
  (ICCE-Asia), pp. 1--4 (2020).
\newblock \doi{10.1109/ICCE-Asia49877.2020.9277040}

\bibitem{KB2019}
Klein, P., Bergmann, R.: Generation of complex data for ai-based predictive
  maintenance research with a physical factory model.
\newblock In: ICINCO (2019)

\bibitem{HJ2018}
Law, H., Deng, J.: Cornernet: Detecting objects as paired keypoints (2018).
\newblock \doi{10.48550/ARXIV.1808.01244}.
\newblock \urlprefix\url{https://arxiv.org/abs/1808.01244}

\bibitem{MZLXX2020}
Li, M., Zhang, Z., Lei, L., Wang, X., Guo, X.: Agricultural greenhouses
  detection in high-resolution satellite images based on convolutional neural
  networks: Comparison of faster r-cnn, yolo v3 and ssd.
\newblock Sensors \textbf{20}(17) (2020).
\newblock \doi{10.3390/s20174938}.
\newblock \urlprefix\url{https://www.mdpi.com/1424-8220/20/17/4938}

\bibitem{LMBBGHPRDZ2014}
Lin, T., Maire, M., Belongie, S.J., Bourdev, L.D., Girshick, R.B., Hays, J.,
  Perona, P., Ramanan, D., Doll{'{a} }r, P., Zitnick, C.L.: Microsoft {COCO:}
  common objects in context.
\newblock CoRR \textbf{abs/1405.0312} (2014).
\newblock \urlprefix\url{http://arxiv.org/abs/1405.0312}

\bibitem{LDGHHB2016}
Lin, T.Y., Dollár, P., Girshick, R., He, K., Hariharan, B., Belongie, S.:
  Feature pyramid networks for object detection (2016).
\newblock \doi{10.48550/ARXIV.1612.03144}.
\newblock \urlprefix\url{https://arxiv.org/abs/1612.03144}

\bibitem{LQQSJ2018}
Liu, S., Qi, L., Qin, H., Shi, J., Jia, J.: Path aggregation network for
  instance segmentation (2018).
\newblock \doi{10.48550/ARXIV.1803.01534}.
\newblock \urlprefix\url{https://arxiv.org/abs/1803.01534}

\bibitem{LAES2016}
Liu, W., Anguelov, D., Erhan, D., Szegedy, C., Reed, S., Fu, C.Y., Berg, A.C.:
  {SSD}: Single shot {MultiBox} detector.
\newblock In: Computer Vision {\textendash} {ECCV} 2016, pp. 21--37. Springer
  International Publishing (2016).
\newblock \doi{10.1007/978-3-319-46448-0_2}

\bibitem{WDDCSCA2015}
Liu, W., Anguelov, D., Erhan, D., Szegedy, C., Reed, S.E., Fu, C., Berg, A.C.:
  {SSD:} single shot multibox detector.
\newblock CoRR \textbf{abs/1512.02325} (2015).
\newblock \urlprefix\url{http://arxiv.org/abs/1512.02325}

\bibitem{L1982}
Lloyd, S.: Least squares quantization in pcm.
\newblock IEEE Transactions on Information Theory \textbf{28}(2), 129--137
  (1982).
\newblock \doi{10.1109/TIT.1982.1056489}

\bibitem{NFNKA2019}
Mehrabi, N., Morstatter, F., Saxena, N., Lerman, K., Galstyan, A.: A survey on
  bias and fairness in machine learning (2019).
\newblock \doi{10.48550/ARXIV.1908.09635}.
\newblock \urlprefix\url{https://arxiv.org/abs/1908.09635}

\bibitem{CMHM2022}
Peng, C., Zhu, M., Ren, H., Emam, M.: Small object detection method based on
  weighted feature fusion and csma attention module.
\newblock Electronics \textbf{11}, 2546 (2022).
\newblock \doi{10.3390/electronics11162546}

\bibitem{YOLOv1}
Redmon, J., Divvala, S.K., Girshick, R.B., Farhadi, A.: You only look once:
  Unified, real-time object detection.
\newblock CoRR \textbf{abs/1506.02640} (2015).
\newblock \urlprefix\url{http://arxiv.org/abs/1506.02640}

\bibitem{YOLOv2}
Redmon, J., Farhadi, A.: {YOLO9000:} better, faster, stronger.
\newblock CoRR \textbf{abs/1612.08242} (2016).
\newblock \urlprefix\url{http://arxiv.org/abs/1612.08242}

\bibitem{YOLOv3}
Redmon, J., Farhadi, A.: Yolov3: An incremental improvement.
\newblock CoRR \textbf{abs/1804.02767} (2018).
\newblock \urlprefix\url{http://arxiv.org/abs/1804.02767}

\bibitem{RHGS2015}
Ren, S., He, K., Girshick, R., Sun, J.: Faster r-cnn: Towards real-time object
  detection with region proposal networks (2015).
\newblock \doi{10.48550/ARXIV.1506.01497}.
\newblock \urlprefix\url{https://arxiv.org/abs/1506.01497}

\bibitem{RTGRS2019}
Rezatofighi, H., Tsoi, N., Gwak, J., Sadeghian, A., Reid, I., Savarese, S.:
  Generalized intersection over union: A metric and a loss for bounding box
  regression (2019).
\newblock \doi{10.48550/ARXIV.1902.09630}.
\newblock \urlprefix\url{https://arxiv.org/abs/1902.09630}

\bibitem{SLJSRAEVR2014}
Szegedy, C., Liu, W., Jia, Y., Sermanet, P., Reed, S., Anguelov, D., Erhan, D.,
  Vanhoucke, V., Rabinovich, A.: Going deeper with convolutions (2014).
\newblock \doi{10.48550/ARXIV.1409.4842}.
\newblock \urlprefix\url{https://arxiv.org/abs/1409.4842}

\bibitem{WLYWCH2019}
Wang, C.Y., Liao, H.Y.M., Yeh, I.H., Wu, Y.H., Chen, P.Y., Hsieh, J.W.: Cspnet:
  A new backbone that can enhance learning capability of cnn (2019).
\newblock \doi{10.48550/ARXIV.1911.11929}.
\newblock \urlprefix\url{https://arxiv.org/abs/1911.11929}

\bibitem{WY2021}
Wang, W., Wang, Y.: Underwater target detection system based on yolo v4.
\newblock In: 2021 2nd International Conference on Artificial Intelligence and
  Information Systems, ICAIIS 2021. Association for Computing Machinery, New
  York, NY, USA (2021).
\newblock \doi{10.1145/3469213.3470310}.
\newblock \urlprefix\url{https://doi.org/10.1145/3469213.3470310}

\bibitem{SSAMSetAl2021}
Zaidi, S.S.A., Ansari, M.S., Aslam, A., Kanwal, N., Asghar, M., Lee, B.: A
  survey of modern deep learning based object detection models (2021).
\newblock \doi{10.48550/ARXIV.2104.11892}.
\newblock \urlprefix\url{https://arxiv.org/abs/2104.11892}

\bibitem{ZR2019}
Zhao, K., Ren, X.: Small aircraft detection in remote sensing images based on
  yolov3.
\newblock IOP Conference Series: Materials Science and Engineering
  \textbf{533}(1), 012,056 (2019).
\newblock \doi{10.1088/1757-899X/533/1/012056}.
\newblock \urlprefix\url{https://dx.doi.org/10.1088/1757-899X/533/1/012056}

\end{thebibliography}
\bibliographystyle{bibtex/spmpsci.bst}
\end{document}